\documentclass[11pt, a4paper, logo, singlecolumn, copyright]{gensyn}

\usepackage{hyperref}
\usepackage{url}
\usepackage{xspace}
\usepackage{mathtools}
\usepackage{algorithm}
\usepackage{algorithmic}
\usepackage{enumitem}
\usepackage{tabularx}
\usepackage{adjustbox}
\usepackage{booktabs}
\usepackage{makecell}
\usepackage[table]{xcolor}
\usepackage{graphicx}
\usepackage{subcaption}

\usepackage{booktabs}
\usepackage{hyperref}
\usepackage{url}
\usepackage{alertmessage}
\usepackage{xspace}
\usepackage{graphicx}
\usepackage{multirow}
\usepackage{rotating}
\usepackage{tikz,lipsum}

\usepackage[colorinlistoftodos,textsize=tiny,textwidth=35pt]{todonotes}
\newcommand{\Comments}{1}
\newcommand{\mynote}[2]{\ifnum\Comments=1\textcolor{#1}{#2}\fi}
\newcommand{\mytodo}[2]{\ifnum\Comments=1\todo[linecolor=#1!80!black,backgroundcolor=#1,bordercolor=#1!80!black]{#2}\fi}
\ifnum\Comments=1
\paperwidth=\dimexpr \paperwidth + 50pt\relax
\oddsidemargin=\dimexpr\oddsidemargin + 25pt\relax
\evensidemargin=\dimexpr\evensidemargin + 25pt\relax
\marginparwidth=\dimexpr \marginparwidth + 25pt\relax
\fi

\usepackage{array}
\newcolumntype{P}{>{\centering\arraybackslash}p{2.5cm}}
\newcolumntype{M}{>{\centering\arraybackslash\footnotesize}m{.78cm}}
\newcolumntype{S}{>{\centering\arraybackslash\tiny}m{2cm}}

\newtcbtheorem[auto counter,number within=section]{obs}%
{Observation}{fonttitle=\bfseries\upshape, fontupper=\slshape,
	arc=0mm, colback=cyan!5!white,colframe=cyan!75!white}{theorem}

\newtcbtheorem[auto counter,number within=section]{ins}%
{Insight}{fonttitle=\bfseries\upshape, fontupper=\slshape,
	arc=0mm, colback=green!5!white,colframe=green!75!white}{theorem}

\usepackage{amsmath,amsfonts,bm}

\def\Figref#1{Figure~\ref{#1}}

\def\Secref#1{Section~\ref{#1}}
\def\eqref#1{equation~\ref{#1}}
\def\Eqref#1{Equation~\ref{#1}}
\def\1{\bm{1}}

\DeclareMathAlphabet{\mathsfit}{\encodingdefault}{\sfdefault}{m}{sl}
\SetMathAlphabet{\mathsfit}{bold}{\encodingdefault}{\sfdefault}{bx}{n}

\newcommand{\R}{\mathbb{R}}

\DeclareMathOperator*{\argmin}{arg\,min}

\title{Training-Free Dynamic Upcycling of Expert Language Models}

\author[1]{Eros Fanì}
\author[1]{Oğuzhan Ersoy}

\affil[1]{Gensyn}

\correspondingauthor{eros.fani@gmail.com, oguzhan@gensyn.ai}

\reportnumber{} % Leave blank if n/a 

\def\Figref#1{Figure~\ref{#1}}

\def\Secref#1{Section~\ref{#1}}
\def\eqref#1{equation~\ref{#1}}
\def\Eqref#1{Equation~\ref{#1}}
\def\Tabref#1{Table~\ref{#1}}

\def\1{\bm{1}}

\definecolor{darkgreen}{rgb}{0,0.7,0}

\newcommand{\clmI}{\textsc{CLMsmall}\xspace}
\newcommand{\clmII}{\textsc{CLMmedium}\xspace}
\newcommand{\reasoningI}{\textsc{Rmedium}\xspace}
\newcommand{\reasoningII}{\textsc{Rlarge}\xspace}

\newcommand\norm[1]{\left\lVert#1\right\rVert}
\newcommand\method{\textsc{DUME}\xspace}
\newcommand\methodplus{\textsc{DUME+}\xspace}
\newcommand\methodfull{Dynamic Upcycling MoE\xspace}
\newcommand{\eg}{\textit{e.g.,}\xspace}
\newcommand{\ie}{\textit{i.e.,}\xspace}

\newcommand{\stdsize}{\fontsize{6}{6}\selectfont}
\newcommand{\num}[2]{#1 \stdsize{$\pm$ #2}}
\newcommand{\bum}[2]{\textbf{#1 \stdsize{$\pm$ #2}}}
\newcommand{\uum}[2]{\underline{#1 \stdsize{$\pm$ #2}}}

\newcommand{\gnum}[2]{\color{darkgreen} #1 \stdsize{$\pm$ #2}}
\newcommand{\gbum}[2]{\color{darkgreen} \textbf{#1 \stdsize{$\pm$ #2}}}
\newcommand{\guum}[2]{\color{darkgreen} \underline{#1 \stdsize{$\pm$ #2}}}

\begin{abstract}
	Large Language Models (LLMs) have achieved remarkable performance on a wide range of specialized tasks, exhibiting strong problem-solving capabilities. However, training these models is prohibitively expensive, and they often lack domain-specific expertise because they rely on general knowledge datasets. Expertise finetuning can address this issue; however, it often leads to overspecialization, and developing a single multi-domain expert remains difficult due to diverging objectives. Furthermore, multitask training is challenging due to interference and catastrophic forgetting. Existing work proposes combining the expertise of dense models within a Mixture of Experts (MoE) architecture, although this approach still requires multitask finetuning. To address these issues, we introduce \methodfull (\method), a novel approach that reuses dense experts trained on different domains to construct a unified MoE model. Our method builds a single multitask model that preserves the capabilities of the original dense experts without requiring additional training. \method is both cost-efficient and scalable: by leveraging the closed-form solution of ridge regression, it eliminates the need for further optimization and enables experts to be added or removed dynamically.
Our extensive experiments on different tasks (in causal language modeling and reasoning settings) with several model sizes (between 115M and 9B parameters) and architectures demonstrate that DUME consistently outperforms baseline approaches. In particular, in a reasoning setting, \method can also surpass the dense expert model's performance in its domain, achieving 104\% of the dense expert's performance on average.
Finally, we also show that the \method model can be fine-tuned to further improve performance in complex reasoning scenarios. Our code is available at: \href{https://github.com/gensyn-ai/dume/}{github.com/gensyn-ai/dume}. 
\end{abstract}

\begin{document}

\maketitle

\begin{abstract}
  Large Language Models (LLMs) have achieved remarkable performance on a wide range of specialized tasks, exhibiting strong problem-solving capabilities. However, training these models is prohibitively expensive, and they often lack domain-specific expertise because they rely on general knowledge datasets. Expertise finetuning can address this issue; however, it often leads to overspecialization, and developing a single multi-domain expert remains difficult due to diverging objectives. Furthermore, multitask training is challenging due to interference and catastrophic forgetting. Existing work proposes combining the expertise of dense models within a Mixture of Experts (MoE) architecture, although this approach still requires multitask finetuning. To address these issues, we introduce \methodfull (\method), a novel approach that reuses dense experts trained on different domains to construct a unified MoE model. Our method builds a single multitask model that preserves the capabilities of the original dense experts without requiring additional training. \method is both cost-efficient and scalable: by leveraging the closed-form solution of ridge regression, it eliminates the need for further optimization and enables experts to be added or removed dynamically.
Our extensive experiments on different tasks (in causal language modeling and reasoning settings) with several model sizes (between 115M and 9B parameters) and architectures demonstrate that DUME consistently outperforms baseline approaches. In particular, in a reasoning setting, \method can also surpass the dense expert model's performance in its domain, achieving 104\% of the dense expert's performance on average.
Finally, we also show that the \method model can be fine-tuned to further improve performance in complex reasoning scenarios. Our code is available at: \href{https://github.com/gensyn-ai/dume/}{github.com/gensyn-ai/dume}. 
\end{abstract}

\section{Introduction}
\label{sec:intro}

Large Language Models (LLMs) have demonstrated remarkable capabilities across a wide range of tasks \citep{radford2019language}, including specialized applications such as code generation \citep{chen2021evaluating}, mathematical problem-solving \citep{cobbe2021training}, and instruction following \citep{zhou2023instruction}. 
In principle, training a single expert model that effectively generalizes across multiple domains would be desirable; however, in practice, this can be challenging due to optimization difficulties \citep{yu2020gradient}, like negative transfer among heterogeneous tasks, catastrophic forgetting \citep{mccloskey1989catastrophic}, or because of the prohibitive computational cost of scaling such models \citep{hoffmann2022training}, and the communication costs or potential privacy challenges \citep{zhang2020privacy} associated with aggregating large, multi-domain datasets.

Recent works on model merging \citep{he2025mergebench} seek to construct unified models with multi-domain and multi-task capabilities by combining specialized experts (that are trained independently). These experts share the same architecture but differ in their fine-tuned parameters, and are merged into a single model typically through linear operations in parameter space \citep{ilharco2022editing, matena2022merging, wang2024localizing, yadav2023ties}, such as simple weight averaging or task arithmetic. This approach is appealing because it avoids retraining from scratch, instead leveraging existing specialized models to produce a multitask expert. However, naïve aggregation can lead to destructive interference \citep{yadav2023ties}, as experts trained on different domains may correspond to distinct loss landscapes; consequently, their parameters may be optimized for incompatible objectives, resulting in merged models that are suboptimal or detrimental across all tasks.

To mitigate interference, another approach is to ensemble these domain experts rather than merge them into a single model; see Branch-Train-Merge (BTM)~\citep{li2022branchtrainmerge}.  BTM ensembling demonstrates strong performance; however, the inference cost increases linearly with the number of experts relative to a single dense model.
Branch-Train-Mix (BTX) \citep{sukhbaatar2024branch} addresses this issue by merging the experts into a single Mixture of Experts (MoE) with sparse routing. 
MoE architectures \citep{shazeer2017outrageously, jacobs1991adaptive,fedus2022switch, dai2024deepseekmoe, jiang2024mixtral, cai2025survey} allow sparse parameter activation, enabling conditional computation by activating only a subset of model parameters for each input, jointly reducing the effective computational cost during inference.
In BTX, the expert models are merged as follows: the non-MLP parameters of the expert models are averaged, whereas the MLP parameters are used as separate MoE experts, which are sparsely routed by a gating network (or router). 
After constructing the combined MoE model, BTX requires multi-task finetuning to calibrate the randomly initialized router.
Such finetuning can be infeasible because of the computational cost and data privacy concerns since it requires to aggregate all domain data into a single centralized node to train the combined MoE model. More details on the related works can be found in Appx.~\ref{app:relatedwork}.

In this paper, we propose \methodfull (\method) that combines multiple dense experts into a single MoE architecture without requiring further finetuning. 
\method obtains the gating networks of the MoE blocks using the closed-form solution of the ridge regression problem \citep{stigler1981gauss}, inspired by the applications in the horizontal \citep{afonin2021towards, fani2025resource, fani2024accelerating} and vertical federated learning \citep{cai2022efficient, huang2022coresets}.
Our extensive experiments demonstrate that no further training is necessary to approach, or even surpass, the performance of the best dense experts for each domain. 
Moreover, even if router fine-tuning is possible, our method remains superior because it provides a substantially better router initialization. 

Our main contributions can be summarized as follows: 
\begin{itemize}[leftmargin=10pt,itemsep=0pt]
    \item We introduce \method, an embarrassingly parallel method that aggregates existing dense experts into a single, multi-domain MoE model, requiring no further training. 
    \item We demonstrate, with experiments in four different settings (two causal language modeling and two reasoning settings) and with four different model sizes and architectures (Llama3 models with 115M, 1B, and 3B parameters, and Gemma2-9B model) \citep{grattafiori2024llama}, that \method outperforms the baselines in both the causal language modeling and the reasoning scenarios.
    \item \method allows adding more experts later in an incremental learning setting without requiring any additional retraining. Moreover, with \method it is also possible to completely remove the contributions of one or more experts as if they were never included.
    \item Finally, when a unified multi-domain dataset is available, our model can be further trained with a Gemma2-9B architecture to outperform the baselines in the large reasoning setting, achieving 109.9\% of the dense expert's performance on average, compared with 86.5\% for the best baseline.
\end{itemize}

\section{Method}
We first briefly explain ridge regression and MoErging procedures used in our method.

\subsection{Ridge Regression}
\label{sec:background}

Let $f(x; W) = W^\top x$ be a linear predictor parameterized by $W \in \R^{H \times C}$, which maps inputs $x \in \R^H$ to targets $y \in \R^C$. For a given dataset $\mathcal{D} = \{ (x_i, y_i) \}_{i=1}^n$, the following ridge regression problem \citep{stigler1981gauss}:
\begin{equation}
    \label{eq:rr-problem}
    \argmin_{W \in \R^{H \times C}} \sum_{i=1}^n\norm{f(x_i; W) - y_i}^2 + \lambda \norm{W}^2,
\end{equation}
controlled by a Tikhonov hyperparameter $\lambda \in \R^+$, admits the closed-form solution:
\begin{equation}
    \label{eq:rr-solution}
    W^* = (X^\top X + \lambda I_H)^{-1} X^\top Y,
\end{equation}
where $X \in \R^{n \times H}$ and $Y \in \R^{n \times C}$ are constructed by stacking the input and target vectors, respectively, and $I_H$ is the identity matrix of size $H$. The term $\lambda \norm{W}^2$ makes the optimization problem strictly convex, guaranteeing a unique solution \citep{shawetaylor2004kernel}. If the targets are one-hot encoding vectors, the problem in \Eqref{eq:rr-problem} can be effectively purposed to solve classification tasks \citep{rifkin2003regularized}. Importantly, thanks to \Eqref{eq:rr-solution}, it is possible to avoid costly, stochastic optimization algorithms to solve \Eqref{eq:rr-problem}.

If the data is sequentially available in $B$ batches of any size, it is possible to sequentially update the $X^\top X$ and $X^\top Y$ matrices as more batches become available, and finally compute the optimal solution in \Eqref{eq:rr-solution} \citep{bjorck1996numerical, fani2023fed3r}: 

\begin{equation}
    \label{eq:rr-seq}
    W^* = \left ( \sum_{i=1}^B X_i^\top X_i + \lambda I_H \right )^{-1} \sum_{i=1}^B X_i^\top Y_i,
\end{equation}

where $X_i \in \R^{n_i \times H}$ and $Y_i \in \R^{n_i \times C}$ are constructed by stacking the input and target vectors within batch $i$, respectively, with $\sum_{i=1}^B n_i = n$. The matrices $X^\top X \in \R^{H \times H}$ and $X^\top Y \in \R^{H \times C}$ are independent of the number of samples per batch, allowing for batches of different sizes.

\subsection{MoErging}
\label{sec:moerging}

Consider a set of \textit{dense experts} $\{f_d\}_{d=1}^D$, sharing a common transformer architecture constituted by an embedding layer, a sequence of $L$ transformer blocks, and the output head. Each transformer block is composed of an attention block and an MLP block. We refer to the $l\text{-th}$ MLP block of the dense expert $f_d$ as the \textit{MoE expert} $\mathcal{E}_{dl}$. We also assume that each dense expert $f_d$ has been trained to solve different language tasks using the local domain dataset $\mathcal{D}_d$ of size $n_d$. 

To construct a unified multi-domain expert MoE model $F$ that leverages the capabilities of pre-trained dense experts $f_d$ while circumventing the need for expensive multi-task retraining, the MoErging approach averages the parameters of all dense expert layers except the MLP layers. The rationale for excluding the MLP layers from averaging is that the non-MLP layers are less specialized than the MLP ones, as noted by \citep{sukhbaatar2024branch}. Our analysis on the similarity of MLP parameters (and non-MLP parameters) for our domain experts can be found in Appx.~\ref{sec:cka}.
The MLP layers are, instead, combined into MoE blocks as $\text{MoE}_l(x) = \sum_{d=1}^D \left [g(x; W_l) \right ]_d \mathcal{E}_{dl}(x)$,
where $\text{MoE}_l$ is the $l\text{-th}$ MoE block of $F$, $W_l$ are the parameters of the router of the $l\text{-th}$ MoE block, and $g$ is a routing function which we define as $g(x; W_l) = \text{Top-k}(\text{Softmax}(W_l^\top x))$, $W_l \in \R^{H \times D}$, where $H$ is the hidden dimension of the dense experts. To have the same forward costs as the dense experts, we set $\text{k}=1$ (see Appx. \ref{sec:hyperparams} for further discussion on the top-k routing of \method).

\subsection{\methodfull (\method)}
\label{sec:dume}

\begin{figure}
    \centering
    \includegraphics[clip, trim=0.0cm 0.0cm 0.0cm 0.0cm,width=\linewidth]{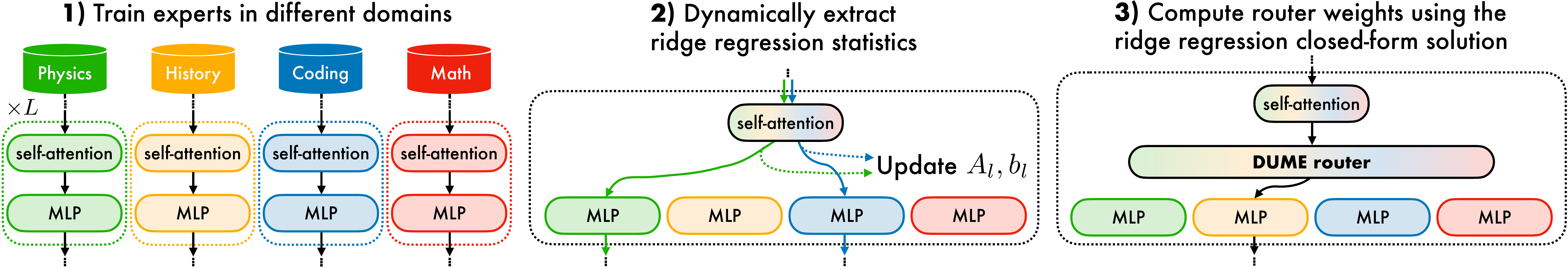}
    \caption{Overview of the \method method. For simplicity and illustration purposes, we only show a single transformer block $l$ among the total $L$ blocks. 
    1) Dense experts are trained on different domains. 2) MoErging phase and ridge regression statistics extraction: the parameters of all the layers but the MLP layers, including the self-attention layers, are averaged. Each input token from all the domains is forwarded to the expert corresponding to its domain label. The feature maps extracted from the layers preceding the MLP blocks are used to update the $A_l$ and $b_l$ matrices. 3) The parameters $W_l^*$ are computed via \Eqref{eq:rr-solution} using the matrices computed at the previous step.}
    \label{fig:method}
\end{figure}

We propose \methodfull (\method), which constructs a unified multi-domain expert MoE model $F$ by exploiting the capabilities of the already-trained dense experts $f_d$ while avoiding costly multi-task re-training. \Figref{fig:method} presents an overview of our method.

After the MoErging phase, \method uses the closed-form ridge regression solution in \Eqref{eq:rr-solution} along with the data's domain labels to initialize the router parameters.
Consider the first MoE block, $\text{MoE}_1$. We initialize two matrices $A_{1,0} = 0_{H \times H}$ and $b_{1,0} = 0_{H \times D}$, where $0_{a \times c}$ is a matrix of zeros of size $a \times c$. The purpose of these two matrices is to iteratively reconstruct the final $X^\top X$ and $X^\top Y$ matrices of \Eqref{eq:rr-solution} (using the notation in \Secref{sec:background}). Given an input text $x_i$ from domain $d$, and the matrix $y_i \in \R^{T_i \times D}$ which is the one-hot encoding matrix associated with the input $x_i$, whose column corresponding to the domain indices is a vector of ones, and all the other values are zeros, \ie $[y_i]_{jk} = 1$ if $k = d$, $0$ if $k \neq d$, the output of all the layers preceding $\text{MoE}_1$, which we call $F_{<1}(x_i)$, has shape $T_i \times H$, where $T_i$ is the number of tokens in $x_i$. To compute the optimal ridge regression weights of the first router, we accumulate the contribution of each new token of an input text $x_i$, in any order, into the $A_1$ and $b_1$ matrices:
\begin{equation}
    \label{eq:update-rule-first}
    \left\{
        \begin{aligned}
            A_{1,t} &= A_{1,t-1} + F_{<1}(x_i)^\top F_{<1}(x_i) \\
            b_{1,t} &= b_{1,t-1} + F_{<1}(x_i)^\top y_i, \\
        \end{aligned}
    \right.
\end{equation}
where the index $t$ indicates the number of recursive updates performed so far. 
We can continue accumulating the contribution of each sample of text in each domain until all the datasets have been forwarded through $F_{<1}$, and use the final $A_1$ and $b_1$ matrices to compute the optimal ridge regression parameters for the first router:
\begin{equation}
    \label{eq:solution-first}
    W_1^* = (A_1 + \lambda I_H)^{-1} b_1.
\end{equation}

If the input text is processed in batches $X_i$ of size $B$, the feature map $F_{<1}(X_i)$ and the stacked one-hot tensor $Y_i$ would both have shape $B \times T_i \times H$\footnote{The number of tokens $T_i$ for each input text within a batch can be made the same by methods like padding.}. In this case, the matrices $A_1$ and $b_1$ can analogously be accumulated by reshaping $F_{<1}(X_i)$ and $Y_i$ to the dimension $BT_i \times H$. Importantly, thanks to the sequential property of ridge regression in \Eqref{eq:rr-seq} \citep{bjorck1996numerical}, the update rule in \ref{eq:update-rule-first} guarantees that the final parameters are exactly the same optimal parameters as if we computed the ridge regression solution using \Eqref{eq:rr-solution}. Finally, to address possible domain unbalanced distributions~\citep{ legate2023guiding},
we normalize $W^*_1$ by dividing each column by its norm: $[W^*_1]_{:,c} = [W^*_1]_{:,c} / \norm{[W^*_1]_{:,c}}$.

In general, all the MoE routers' parameters and the final optimal weights can be computed analogously. However, a layer $l$ would require all the MoE routers $l>1$ to be initialized beforehand to compute $F_{<l}(x_i)$, \ie the output of all the layers preceding $\text{MoE}_l$. A possible solution is to compute the optimal weights for the routers sequentially, starting with router $1$ and proceeding to router $L$. However, this strategy would require increasing the number of forward passes by a factor of $L$, since one would need to forward the entire datasets from all domains every time for each new router. To avoid this issue, we instead forward the text from all the domains datasets only once, update all the matrices $A_{l, t}$ and $b_{l, t}$ simultaneously and, instead of using the gating layer to forward each feature map to the MoE experts, we instead deterministically forward each feature map to the MoE expert corresponding to the domain of the input text. The general update rule of the matrices $A_{l, t}$ and $b_{l, t}$ is:
\begin{equation}
    \label{eq:update-rule-general}
    \left\{
        \begin{aligned}
            A_{l,t} &= A_{l,t-1} + F_{d,<l}(x_i)^\top F_{d,<l}(x_i) \\
            b_{l,t} &= b_{l,t-1} + F_{d,<l}(x_i)^\top y_i, \\
        \end{aligned}
    \right.
    \Longrightarrow\;
    W_l^* = (A_l + \lambda I_H)^{-1} b_l,
\end{equation}

where $A_{l,0} = 0_{H \times H}$, $b_{l,0} = 0_{H \times D}$, $F_{d,<l}(x_i)$ indicates the feature maps extracted with this strategy for domain $d$. In particular, note that, for $d=1$ and $\forall l$, $F_{d,<l}(x_i) = F_{<l}(x_i)$. Finally, the final weights of each router are normalized: $[W^*_l]_{:,c} = [W^*_l]_{:,c} / \norm{[W^*_l]_{:,c}}$. In \Secref{sec:dumevariants}, as an alternative, we also analyse the solution computing the optimal weights for the routers sequentially.

For the MoE routers $l>1$, the computed parameters are not optimal in the sense of ridge regression, because the input space used to compute the ridge regression solution depends on the domain of the input text. In other words, given a router $l>1$, each input of the domain dataset $\mathcal{D}_d$ is forwarded to the MoE experts $\mathcal{E}_{dl'}$ for all the preceding MoE blocks $l'<l$, which constitutes a different model architecture from the case of domain $d' \neq d$, for which the MoE experts choice is $\mathcal{E}_{d'l'}$ for all $l'<l$. We refer to this problem as \emph{domain-conditioned feature shift}. However, we argue that the effects of having different input spaces on the final solution are negligible, because \textit{a)} the latent spaces would still have the same dimensionalities, and \textit{b)} the self-attention layers, which are the same for all the domains as their parameters are averaged from the original dense experts, are structurally well-suited to encode relational and syntax knowledge which prevails over the domain knowledge in shaping the distribution of the latent feature maps. Indeed, in \Secref{sec:experiments}, we show that our approximate ridge regression solution already outperforms existing methods. Moreover, in the following section, we introduce \method REC, a variant of \method which preserves the optimality of the ridge regression solution at the expense of more time and computational costs.
The complete pseudo-code of \method can be found in Appx.~\ref{sec:pseudocode}, and the computational costs are discussed in Appx.~\ref{sec:compcosts}.

In our method, the datasets can be accessed sequentially, as the ridge regression solution is schedule-invariant \citep{fani2025resource, wang2022schedule}. In other words, if the domain datasets are made available sequentially, \method enables continual learning, as it can dynamically update the routers without incurring catastrophic forgetting \citep{mccloskey1989catastrophic}. 

\subsection{\method variants}
\label{sec:dumevariants}

Note that, although routers initialized with our \method are already sufficient to surpass the baselines (as we show in \Secref{sec:experiments}), they can be further finetuned. We refer to our method with additional router finetuning as \methodplus. 
Moreover, in Appx. \ref{sec:ood}, we show that our method generalizes even when the distribution of the datasets used to update the $A_l$ and $b_l$ matrices differs from the test distribution. We refer to this variation as \method out-of-distribution (\method OOD).

Finally, to check if the \emph{domain-conditioned feature shift} problem introduced in \Secref{sec:dume}, making the input space used to compute the ridge regression solution independent from the domain of the input text, we introduce Recursive \method (\method REC). Unlike \method, \method REC constructs the routers one at a time, starting with the first and ending with the last. The construction of the weights $W^*_1$ is equivalent to \method. However, for $l\geq2$, the same feature extractor $F_{<l}$ using the pre-computed routers $g_{l'}$, $\forall l'<l$ is used, independently from the domain of the input tokens. In Appx. \ref{sec:dumerec} we show that the benefits of \method REC over \method are negligible, and that the feature maps of the two models remain highly aligned at all depths (cosine similarity of at least 0.997 between their principal-component projections), i.e., the domain-conditioned feature shift is small in practice.

\subsection{Dynamic addition and removal of experts}
\label{sec:dumedyn}

\begin{figure}[t]
    \centering
    \begin{minipage}{0.47\linewidth}
        \centering
        \includegraphics[width=\linewidth]{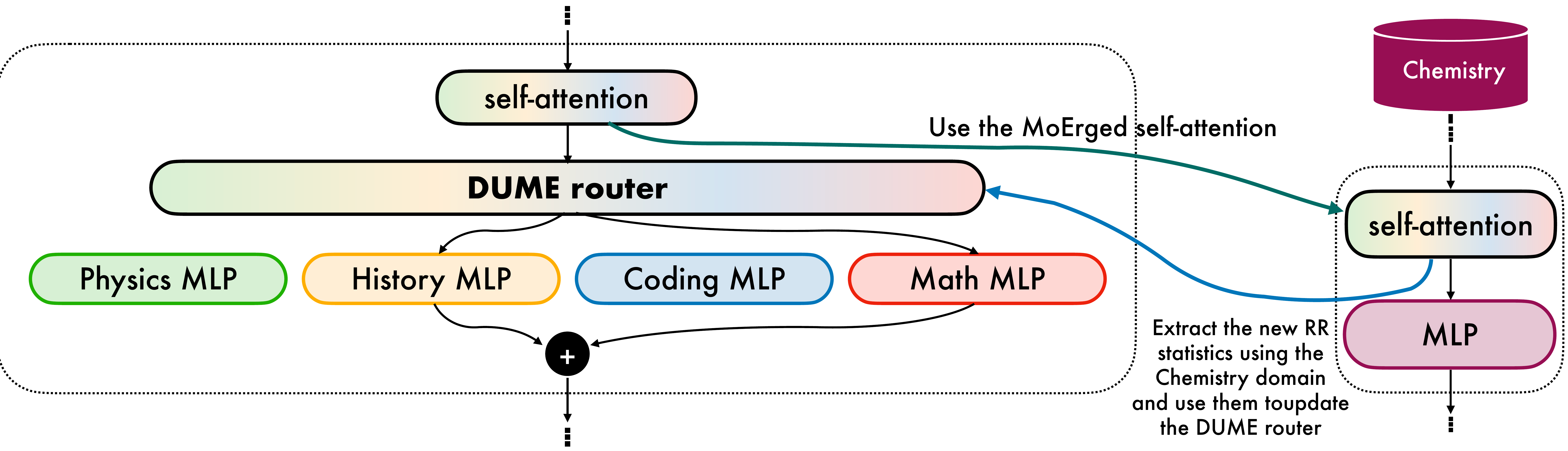}
        \subcaption{}
        \label{fig:dumeincra}
    \end{minipage}
    \hfill
    \begin{minipage}{0.47\linewidth}
        \centering
        \includegraphics[width=\linewidth]{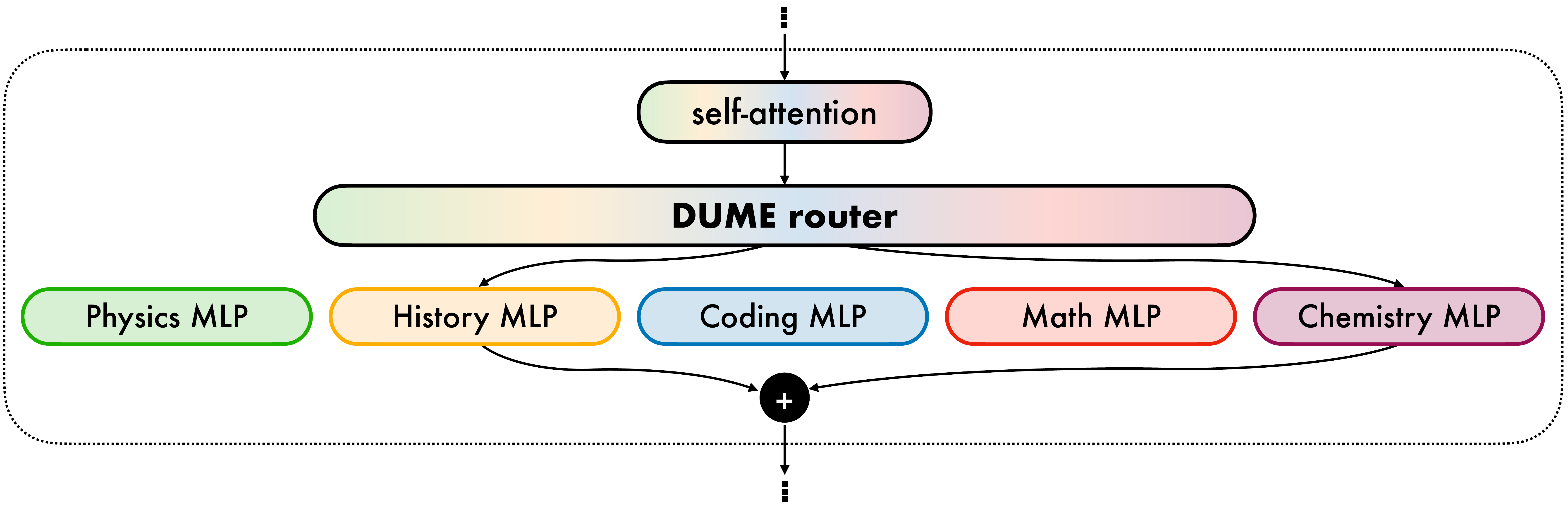}
        \subcaption{}
        \label{fig:dumeincrb}
    \end{minipage}
    \hfill
    \caption{\method allows dynamically MoErging new experts without retraining the router from scratch or with continual learning. The non-MLP parameters can be updated in the MoErged model using a moving average, or the existing parameters can be used. Then, (a) the self-attention block can be used to extract ridge regression statistics for the new Chemistry domain and (b) the Chemistry MLP experts can be added to the pool of MoE experts.}
    \label{fig:dumeincr}
\end{figure}

\method is inherently dynamic, enabling the seamless addition or removal of experts as expertise requirements evolve. Specifically, the \method router can efficiently incorporate or discard domain knowledge in the MoErged \method router by updating the \method statistics matrices. In practice, given a new domain $d_{\text{new}}$, the contributions $F_{d,<l}(x_i)^\top F_{d_{\text{new}},<l}(x_i)$ and $F_{d_{\text{new}},<l}(x_i)^\top y_i$ can be added to or removed from the matrices $A_l$ and $b_l$, and then the router parameters can be recalculated using the closed form solution in \Cref{eq:rr-solution}. An illustration of this process is provided in \Figref{fig:dumeincr}.

The updating of the non-MLP parameters and the exact parameters to be used to extract the new statistics can be handled in different ways, as we show in \Secref{sec:incrementalmain}. Importantly, when the non-MLP parameters are aggregated using reversible merging strategies, such as simple averaging, the contribution of an undesired expert can be removed exactly, enabling perfect machine unlearning~\citep{unlearning2021}. In particular, router contributions can be either stored explicitly or recomputed when needed, while non-MLP parameters can be updated analytically. For example, if the expert parameters are merged through model averaging $\Bar{\theta} = \frac{1}{D} \sum_d \theta_d$, then removing the contribution of expert $\theta_{d_{new}}$ yields the updated parameters $\Bar{\theta}_{\text{new}} = \frac{1}{D-1}(D \Bar{\theta}_{\text{new}} - \theta_{d_{\text{new}}})$.

\section{Experiments}
\label{sec:experiments}

\subsection{Experimental Setup}
\label{sec:details}

All our experiments are conducted with one to three replicas, in four settings: two \emph{Causal Language Modeling} settings (\clmI and \clmII), in which the task across all domain datasets is next-token prediction, and two \emph{Reasoning} settings (\reasoningI and \reasoningII), in which each domain has its own distinct reasoning task. We run our experiments on up to 4 NVIDIA H100 GPUs with 80 GB of HBM3 memory. Below, we present our implementation choices and default hyperparameters. All of our default hyperparameter values are the result of our hyperparameter tuning. Further details are provided in Appx. \ref{sec: exp_setup_appx}, \ref{sec:merging} and \ref{sec:hyperparams}.

\paragraph{Expert Models and Domain Datasets.} In both the CLM settings, we train our own experts for the next token prediction task, starting from the same seed model across the experts. For \clmI experiments, we use the model and the domain datasets from \citep{ersoy2025hdee}. First, we pre-trained a seed 115M Llama3 model \citep{grattafiori2024llama} on the OpenWebText corpus dataset \citep{Gokaslan2019OpenWeb}, and finetuned expert models using M2D2 domains \citep{reid2022m2d2} (specifically, coding, mathematics, physics, history and events, and philosophy and thinking). For \clmII experiments, we finetuned expert models using a 1B Llama3 base model on the following domains: mathematics (OpenWebMath \citep{paster2023openwebmath}),  biology (peS2o \citep{peS2o}), legal (Pile of Law \citep{hendersonkrass2022pileoflaw}), and dialogue (UltraChat 200k \citep{ding2023enhancing}). Details on expert training for both CLM setups can be found in Appx.~\ref{sec: exp_setup_appx}.

For the \reasoningI experiments, we use four domain-specific LLama3-3B experts from MergeBench \citep{he2025mergebench}, for the following domains: coding, mathematics, multilingual understanding, and instruction following. We perform evaluation on the following datasets: HumanEval \citep{chen2021evaluating} for coding, GSM8k \citep{cobbe2021training} for mathematics, M\_ARC \citep{lai2023okapi} for multilingual understanding, and IFEval \citep{zhou2023instruction} for instruction following. Details on the evaluation metrics for these datasets can be found in Appx.~\ref{sec: exp_setup_appx}. 
To test different model sizes and architectures, we replicate the same experiments with \reasoningII using Gemma2-9B experts \citep{team2024gemma} finetuned on the same domains (again by MergeBench \citep{he2025mergebench}).

\paragraph{Baselines.}
In all our proposed scenarios, we compare our \method and \methodplus methods with the following baselines:
\begin{itemize}[leftmargin=10pt,itemsep=0pt]
    \item \emph{``Domain'' expert} is the dense expert trained on the ``domain'' dataset. We propose this baseline for all the domain dense experts in each setting.
    \item \emph{Model averaging} is the model obtained by averaging the dense experts.
    \item \emph{Oracle} is a MoE model initialized with model MoErging, where the tokens are forwarded to the MoE experts corresponding to the batch domain label. Oracle requires domain labels at test time, which may not be available in practice. 
    The routing could be overly rigid and may not effectively differentiate among individual tokens; yet, this could constitute an advantage in forwarding the tokens to the most appropriate expert, as our experiments also show.
    \item \emph{Random routing} is another MoErged baseline, and unlike Oracle, this baseline includes routers, but they are initialized at random and not trained. Therefore, each token is randomly forwarded to one of the domain MoE experts. This baseline is included as a lower-bound.
    \item \emph{BTX} is proposed by \cite{sukhbaatar2024branch} where the router of MoE is finetuned after MoErging.
\end{itemize}

Additionally, in both the CLM scenarios, given that we trained our dense experts and as such we have access to the task vectors \citep{ilharco2022editing}, we compare \method and \methodplus with additional model merging baselines: \emph{Task arithmetic} \citep{ilharco2022editing, zhou2024metagpt}, \emph{TIES Merging} \citep{yadav2023ties}, \emph{DARE} \citep{yu2024language}, \emph{Fisher Merging} \citep{matena2022merging},  and \emph{Consensus Merging} \citep{wang2024localizing}. More details on these algorithms and their hyperparameter choices are provided in Appx. \ref{sec:merging}.

Moreover, in Appx.~\ref{sec:dumerec}, we compare \method with its \method REC version, and in Appx.~\ref{sec:ood} we also provide the \method OOD baseline (see \Secref{sec:dumevariants} for an introduction to these variants), and in Appx.~\ref{sec:dumevssoftmax} we compare \method with the direct training of a softmax router. Finally, Appx. \ref{sec:dumefrozen} shows the results of \method and \methodplus when not all the parameters of the original dense experts have been trained, but instead only the MLP layers have been trained, and the rest is frozen.

\paragraph{Metrics.} The CLM experiments have been evaluated using the normalized average perplexity score on all five domains. In particular, let $p_d$ be the perplexity score of a certain experiment on domain $d$, and let $\Hat{p}_d$ be the perplexity score of the dense expert $\mathcal{E}_d$ on domain $d$, which is expected to achieve the best performance (lowest perplexity) on domain $d$ among all the other models. We define the normalized average perplexity score $\Bar{p} \in \R^+$ for a given experiment as $\Bar{p} = \frac{100}{D} \sum_{d=1}^D \frac{\Hat{p}_d}{p_d}$. Similarly, we evaluate the reasoning experiments using the normalized average performance score $\Bar{p} \in \R^+$, defined as $\Bar{p} = \frac{100}{D} \sum_{d=1}^D \frac{p_d}{\Hat{p}_d}$, where, in this case, we inverted the fraction in the sum since the best performance corresponds to the highest values for all the metrics of the reasoning datasets. Note that, with these definitions,\emph{it is possible (and desirable) that the value of the metrics for a particular domain exceeds 100}, in which case it means that the experiment surpasses the dense expert in the corresponding domain.

In the following sections we present our CLM and reasoning experiments. In all the tables, the best and second best results (excluding the results of the experts in the corresponding domains, for which the score is 100.0 by definition, and the Oracle baseline, which, unlike the other methods, uses the domain label at test time) are highlighted in \textbf{bold} and \underline{underlined}, respectively. Additionally, results that surpass the Oracle, excluding the dense expert results evaluated on their own domains, are highlighted in {\color{darkgreen} green}.

\subsection{Causal Language Modeling results}
\label{sec:clmres}

\begin{table}[t]
    \caption{\clmI experiments.}
    \label{tab:clm}
    \centering
    \begin{adjustbox}{width=.75\linewidth}
        \begin{tabular}{c|ccccc|c}
            \toprule
            \textbf{Method} & \textbf{Coding} & \textbf{Mathematics} & \textbf{Physics} & \textbf{History} & \textbf{Philosophy} & \textbf{Average} \\
            \midrule
            Coding expert     & \num{100.0}{0.0} & \num{75.1}{2.3}  & \num{83.6}{2.3}  & \num{84.0}{1.9}  & \num{84.6}{1.5}  & \num{85.4}{1.6} \\
            Math expert       & \num{92.9}{1.5}  & \num{100.0}{0.0} & \num{83.7}{2.1}  & \num{78.4}{2.8}  & \num{79.2}{2.4}  & \num{86.8}{1.8} \\
            Physics expert    & \num{85.4}{1.8}  & \num{67.3}{2.1}  & \num{100.0}{0.0} & \num{84.4}{1.6}  & \num{84.9}{1.2}  & \num{84.4}{1.3} \\
            History expert    & \num{72.1}{3.1}  & \num{47.3}{3.6}  & \num{72.4}{3.2}  & \num{100.0}{0.0} & \num{98.6}{0.2}  & \num{78.1}{2.0} \\
            Philosophy expert & \num{73.8}{2.5}  & \num{49.6}{3.1}  & \num{73.5}{3.0}  & \num{97.0}{0.5}  & \num{100.0}{0.0} & \num{78.8}{1.8} \\
            \midrule
            Model averaging   & \num{90.5}{3.7}  & \num{69.2}{6.2}  & \num{86.2}{3.5}  & \num{92.5}{0.9}  & \num{94.2}{0.7}  & \num{86.5}{2.9} \\
            Task Arithmetic   & \uum{98.5}{2.0}  & \num{84.8}{1.0}  & \num{92.7}{2.3}  & \num{91.6}{1.3}  & \num{93.1}{1.3}  & \num{92.2}{1.6} \\
            TIES Merging      & \gbum{100.7}{1.7} & \num{86.6}{0.7}  & \num{94.0}{2.2}  & \num{91.0}{1.5}  & \num{92.7}{1.6}  & \num{93.0}{1.5} \\
            DARE              & \uum{98.5}{2.0}  & \num{84.7}{1.0}  & \num{92.7}{2.3}  & \num{91.6}{1.3}  & \num{93.1}{1.3}  & \num{92.1}{1.6} \\
            Fisher Merging    & \num{95.6}{3.9}  & \num{75.9}{5.9}  & \num{90.7}{3.8}  & \num{94.9}{1.1}  & \num{96.4}{0.9}  & \num{90.7}{3.1} \\
            Consensus Merging & \uum{98.5}{4.0}  & \num{80.7}{6.3}  & \num{92.7}{3.9}  & \num{83.6}{1.0}  & \num{85.2}{0.8}  & \num{88.1}{3.2} \\
            \midrule
            Oracle            & \num{98.7}{2.2}  & \num{96.6}{5.9}  & \num{98.4}{2.8}  & \num{99.2}{1.4}  & \num{99.4}{1.1}  & \num{98.4}{2.7} \\
            Random routing    & \num{88.5}{0.8}  & \num{69.1}{2.4}  & \num{86.3}{1.6}  & \num{91.9}{0.8}  & \num{93.1}{0.8}  & \num{85.8}{1.2} \\
            BTX               & \num{96.4}{1.8}  & \uum{94.4}{5.3}  & \uum{95.6}{2.5}  & \num{98.1}{1.3}  & \uum{98.4}{1.0}  & \uum{96.6}{2.4} \\
            \rowcolor{gray!15}
            \textbf{\method (ours)}  & \num{97.5}{2.0}  & \num{92.3}{5.1}  & \bum{95.8}{2.4}  & \bum{98.9}{1.3}  & \bum{98.9}{0.9}  & \bum{96.7}{2.3} \\
            \rowcolor{gray!15}
            \textbf{\methodplus (ours)}  & \num{96.5}{1.8}  & \bum{95.0}{5.5}  & \uum{95.6}{2.7}  & \uum{98.2}{1.4}  & \uum{98.4}{1.0}  & \bum{96.7}{2.5} \\
            \bottomrule
        \end{tabular}
    \end{adjustbox}
\end{table}

\paragraph{\clmI} \Tabref{tab:clm} presents the results for the \clmI experiments. As can be observed, both \method and \methodplus surpass all baselines in the average normalized score, approaching the Oracle, without requiring domain labels at test time. In particular, \method is already sufficient to surpass the best of our comparison methods, BTX, in four of five domains and to slightly surpass it in the average score, without requiring any additional training. Moreover, \methodplus does not improve the performance of \method, indicating that \method is already sufficient to achieve near-optimal performance, which is likely very difficult to further improve, also given the simplicity of this setting. Indeed, in \Secref{sec:dumereasoning} we show that this is not always true, since in more complex scenarios with larger models, there is a clear benefit to further training the routers. Finally, we find that TIES Merging performs exceptionally well in this setting, approaching the performance of the MoErging baselines. In Appx. \ref{sec:mergingmethods} we further investigate what happens when combining \method with other merging algorithms in both the \clmI and \clmII settings, and find that \method with TIES merging achieves even better performance.

\begin{table}[t]
    \caption{\clmII experiments.}
    \label{tab:clm2}
    \centering
    \begin{adjustbox}{width=.75\linewidth}
        \begin{tabular}{c|cccc|c}
            \toprule
            \textbf{Method} & \textbf{Mathematics} & \textbf{Biology} & \textbf{Legal} & \textbf{Dialogue} & \textbf{Average} \\
            \midrule
            Math expert        & \num{100.0}{0.0} & \num{91.4}{0.1} & \num{41.6}{0.1} & \num{91.1}{1.8} & \num{81.0}{0.4} \\
            Biology expert     & \num{71.2}{0.2}  & \num{100.0}{0.0} & \num{31.6}{1.1} & \num{69.9}{7.3} & \num{68.2}{2.1} \\
            Legal expert       & \num{71.0}{0.2}  & \num{86.4}{0.1}  & \num{100.0}{0.0} & \num{89.5}{1.7} & \num{86.7}{0.4} \\
            Dialogue expert    & \num{66.4}{0.2}  & \num{88.0}{0.1}  & \num{33.2}{0.1} & \num{100.0}{0.0} & \num{71.9}{0.1} \\
            \midrule
            Model averaging    & \num{79.9}{0.8}  & \num{93.5}{2.0}  & \num{50.0}{0.6}  & \num{93.5}{1.9}  & \num{79.2}{0.7} \\
            Task Arithmetic    & \num{83.4}{0.0}  & \num{93.9}{0.0}  & \num{54.8}{0.1}  & \num{94.4}{1.8}  & \num{81.6}{0.5} \\
            TIES Merging       & \num{85.0}{0.9}  & \num{95.8}{1.9}  & \num{56.0}{0.7}  & \num{95.3}{2.1}  & \num{83.0}{0.8} \\
            DARE               & \num{83.1}{0.8}  & \num{93.8}{2.0}  & \num{54.7}{0.6}  & \num{94.4}{1.8}  & \num{81.5}{0.7} \\
            Fisher Merging     & \num{81.2}{0.8}  & \num{92.3}{1.9}  & \num{59.1}{0.7}  & \num{91.4}{1.7}  & \num{81.0}{0.7} \\
            Consensus Merging  & \num{86.8}{1.0}  & \num{96.2}{1.9}  & \num{56.4}{0.8}  & \num{94.9}{1.8}  & \num{83.6}{0.8} \\
            \midrule
            Oracle             & \num{93.4}{0.6}  & \num{97.3}{1.2}  & \num{86.1}{0.9} & \num{97.9}{2.1} & \num{93.7}{0.7} \\
            Random routing     & \num{80.6}{0.7}  & \num{94.8}{0.9}  & \num{47.9}{0.1} & \num{92.9}{2.1} & \num{79.1}{0.8} \\
            BTX                & \bum{92.7}{1.3}  & \num{95.1}{0.7}  & \num{84.9}{1.4} & \num{95.6}{2.0} & \num{92.1}{0.4} \\
            \rowcolor{gray!15}
            \textbf{\method (ours)}     & \uum{92.4}{1.6}  & \uum{96.3}{1.5}  & \gbum{86.4}{0.3} & \uum{95.8}{2.1} & \bum{92.7}{0.8} \\
            \rowcolor{gray!15}
            \textbf{\methodplus (ours)} & \num{91.4}{1.3}  & \bum{96.5}{2.0}  & \uum{84.3}{0.5} & \bum{97.2}{1.9} & \uum{92.4}{0.7} \\
            \bottomrule
        \end{tabular}
    \end{adjustbox}

\end{table}

\paragraph{\clmII} \Tabref{tab:clm2} shows the results in the \clmII setting. Also in this setting, our methods both surpass all the baselines in almost all domains (with the exception of Mathematics, where \method is the second-best, with a small difference of only 0.3\% relative to BTX), this time with a larger margin than in the \clmI setting. Also in this scenario, \methodplus does not surpass \method, confirming the quality of our training-free approach. Interestingly, \method surpasses both \methodplus and the Oracle in the Legal domain. As anticipated in \Secref{sec:details}, we believe that this happens when the advantage of a flexible routing mechanism, which can dynamically forward tokens belonging to the same sample also to experts of different domains, surpasses the advantage of using the expert of the sample domain for all the tokens in that sample in a static way.

\subsection{Reasoning experiments}
\label{sec:dumereasoning}

\begin{table}[t]
    \caption{\reasoningI experiments.}
    \label{tab:reasoning}
    \centering
    \begin{adjustbox}{width=.75\linewidth}
        \begin{tabular}{c|cccc|c}
            \toprule
            \textbf{Method} & \textbf{Mathematics} & \textbf{Multilingual} & \textbf{Coding} & \textbf{Instr. following} & \textbf{Average} \\
            \midrule
            Math expert             & \num{100.0}{0.0} & \num{95.9}{0.0} & \num{90.1}{10.1} & \num{45.3}{1.3} & \num{82.8}{2.8} \\
            Multilingual expert     & \num{5.8}{0.6} & \num{100.0}{0.0} & \num{82.6}{8.2} & \num{20.9}{0.5} & \num{52.3}{1.9} \\
            Coding expert           & \num{11.9}{0.1} & \num{98.3}{1.6} & \num{100.0}{0.0} & \num{50.6}{5.0} & \num{65.2}{1.3} \\
            Instr. following expert & \num{28.6}{0.5} & \num{99.8}{0.0} & \num{84.3}{4.4} & \num{100.0}{0.0} & \num{78.2}{1.0} \\
            Model averaging         & \num{45.8}{1.2} & \num{100.1}{0.0} & \num{102.3}{16.2} & \num{57.2}{3.3} & \num{76.3}{4.9} \\
            \midrule
            Oracle                  & \num{97.0}{2.5} & \num{100.2}{0.0} & \num{97.9}{20.4} & \num{84.0}{1.6} & \num{94.8}{4.9} \\
            Random routing          & \num{38.7}{2.1} & \num{99.8}{0.0} & \num{85.5}{13.1} & \num{58.5}{1.3} & \num{70.6}{3.5} \\
            BTX                     & \num{43.7}{0.5} & \num{99.8}{0.0} & \gnum{105.1}{9.2} & \num{59.0}{0.6} & \num{76.9}{2.5} \\
            \rowcolor{gray!15}
            \textbf{\method (ours)} & \bum{92.0}{0.9} & \gbum{100.4}{0.0} & \guum{108.1}{6.2} & \gbum{86.9}{0.7} & \gbum{96.8}{1.3} \\
            \rowcolor{gray!15}
            \textbf{\methodplus (ours)} & \uum{88.2}{1.2} & \guum{100.2}{0.0} & \gbum{108.8}{10.9} & \uum{83.3}{1.5} & \guum{95.1}{3.0} \\
            \bottomrule
        \end{tabular}
    \end{adjustbox}
\end{table}

\paragraph{\reasoningI} \Tabref{tab:reasoning} shows the results in the \reasoningI setting. For all the domains, both \method and \methodplus, surpass the baselines, and sometimes even surpassing the Oracle. Also in this case, \methodplus falls slightly behind \method, likely for the same motivation discussed for \clmI and \clmII.

\begin{table}[t]
    \caption{\reasoningII experiments.}
    \label{tab:reasoning2}
    \centering
    \begin{adjustbox}{width=.75\linewidth}
        \begin{tabular}{c|cccc|c}
            \toprule
            \textbf{Method} & \textbf{Mathematics} & \textbf{Multilingual} & \textbf{Coding} & \textbf{Instr. following} & \textbf{Average} \\
            \midrule
                Math expert             & 100.0 & 67.1 & 1.1  & 25.2 & 48.4 \\
                Multilingual expert     & 24.5  & 100.0& 70.3 & 28.2 & 55.8 \\
                Coding expert           & 62.5  & 82.5 & 100.0& 46.4 & 72.9 \\
                Instr. following expert & 84.7  & 96.3 & 65.9 & 100.0& 86.7 \\
                Model averaging         & {\color{darkgreen}115.7} & 89.9 & 72.5 & 46.2 & 81.1 \\
            \midrule
                Oracle                  & 105.7 & 97.9 & 104.4& 88.2 & 99.1 \\
                Random routing          & {\color{darkgreen}109.1} & 89.2 & 79.1 & 46.4 & 81.0 \\
                BTX                     & {\color{darkgreen}\num{121.4}{5.5}} & \num{93.8}{1.4} & \num{91.5}{12.5} & \num{43.7}{4.8} & \num{87.6}{6.0} \\
                \rowcolor{gray!15}
                \textbf{\method (ours)} & {\color{darkgreen}\underline{130.4}} & \underline{96.5} & {\color{darkgreen}\underline{107.7}} & \textbf{81.6} & {\color{darkgreen}\underline{104.0}} \\
                \rowcolor{gray!15}
                \textbf{\methodplus (ours)} & {\color{darkgreen}\bum{148.1}{15.4}} & {\color{darkgreen}\bum{98.5}{1.7}} & {\color{darkgreen}\bum{112.2}{4.0}} & \uum{80.6}{2.1} & {\color{darkgreen}\bum{109.8}{5.8}} \\
            \bottomrule
        \end{tabular}
    \end{adjustbox}
\end{table}

\paragraph{\reasoningII} \Tabref{tab:reasoning2} shows the experiments for the \reasoningII setting. In this scenario, we demonstrate that our \method and \methodplus methods work with other, larger architectures (as we use a Gemma2-9B model, unlike the other settings where we used Llama3 models of various sizes). Here, our methods surpass the baselines by a large margin. Interestingly, unlike the other settings, here the additional training of the router in \methodplus allows achieving better final performance (109.8 for \methodplus vs. 104.0 for \method). We argue that this is most likely due to the much higher number of transformer blocks (42 for Gemma2-9B vs. 24 for LLama3-3B). Indeed, when extracting the \method statistics, as discussed in \Secref{sec:dume}, the computed router parameters are not optimal in the sense of ridge regression, and the error increases the deeper the router is in the model. In any case, \methodplus yields a valid improvement over the final results and largely surpasses BTX, which is trained for the same number of training steps. Finally, we observe that the Math expert is performing poorly because it is overfitted to the finetuning dataset, we provide additional comments and examples on this in Appx.~\ref{sec:math}.

\subsection{Dynamic addition of new experts}
\label{sec:incrementalmain}

\begin{figure}[t]
    \centering
    \includegraphics[width=\linewidth]{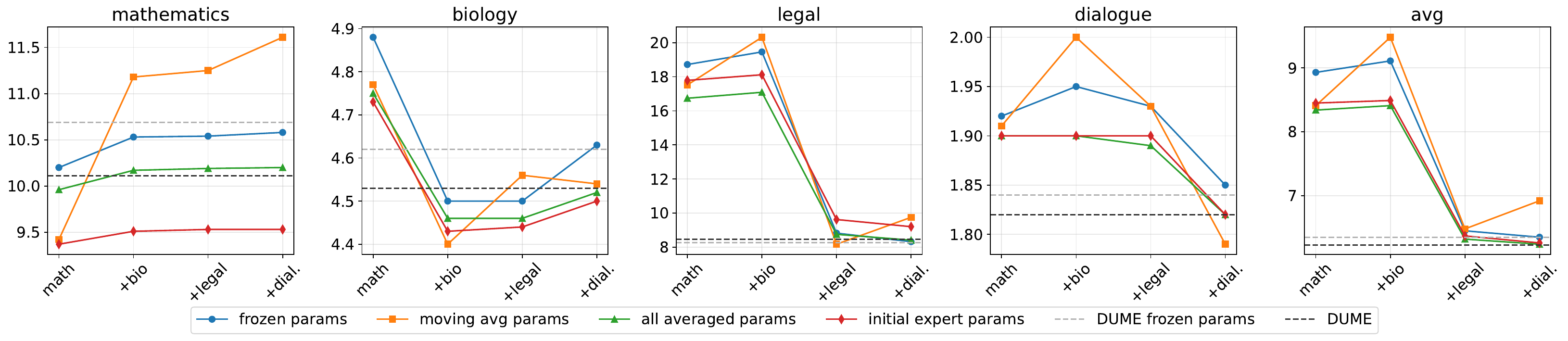}
    \caption{Dynamic incremental addition of new experts with \method in the \clmII setting.}
    \label{fig:incrclm2}
\end{figure}

\Figref{fig:incrclm2} shows experiments in the \clmII setting where the experts are added one at a time, as described in \Secref{sec:dumedyn}. Each one of the five plots shows the (non-normalized) perplexity score on the domain dataset in the tile, with the final plot showing the average perplexity across the four \clmII domains. The x-axis indicates how \method behaves when adding experts sequentially, starting with the mathematics expert. The different curves show different settings and strategies on how to handle the dynamic MoErging of the new experts:
\begin{itemize}[leftmargin=10pt,itemsep=0pt]
    \item In the \emph{frozen params} strategy, the original dense experts have trained only the MLP layers, while the other parameters are frozen (meaning that all the domain experts share the same non-MLP parameters, which are the same as the non-MLP parameters of the seed model). Therefore, the \method statistics are extracted using the same seed model parameters for every addition.
    \item The \emph{moving avg params} strategy updates the non-MLP parameters with a moving average. The updated parameters are used in the MoErged model to extract the new \method statistics for each new expert.
    \item The \emph{all averaged params} strategy is an oracle method that knows the final average non-MLP parameters in advance, and uses these same parameters for the extraction of the \method statistics for each newly added expert.
    \item The \emph{initial expert params} strategy does not update the non-MLP parameters, originally initialized with the ones of the mathematics expert.
    \item \emph{DUME} is the DUME perplexity on all four domains, without sequential MoErging of new experts (all the domain experts are available from the beginning).
    \item \emph{DUME frozen params} is similar to the previous one, but here we use the same domain experts as in \emph{frozen params}.
\end{itemize}

Across all plots, we observe that each time a new domain expert is added, the corresponding domain's perplexity decreases, and this holds for all four incremental \method strategies and all domains. As expected, the final performance of \emph{frozen params} matches the \emph{DUME frozen params} one (within a certain tolerance due to the intrinsic randomness of GPU operations). Indeed, the \method statistics are extracted using the same model parameters in both cases, and the two methods are mathematically equivalent.

All the strategies perform similarly, with the sole exception of \emph{moving avg params}, whose performances are slightly more unstable, likely due to extracting \method statistics from different input feature spaces for each domain due to the moving average of the non-MLP parameters. Interestingly, we also observe that the perplexity scores for the \emph{initial expert params} and \emph{all averaged params} are similar, with the latter only slightly better than the former. This indicates how good the non-MLP parameters of the dense experts already are, and when comparing with the \emph{moving avg params}, it also underscores the importance of extracting \method statistics from nearly static input feature spaces.

In Appx. \ref{sec:moreincr} we provide additional experiments with dynamic addition of new experts. Moreover, in Appx. \ref{sec:decr} we also empirically show that \method allows for dynamical removal of experts from existing \method MoErged models.

\section{Conclusion}

In this work, we presented \method, a new embarrassingly parallel method that upcycles dense experts trained on different domains to construct a unified Mixture-of-Experts model. Contrary to existing methods, \method does not require any additional multitask or multidomain training, as it initializes all final MoE parameters at minimal cost by leveraging the sequential properties of ridge regression and simple model averaging. In addition to surpassing existing baselines, \method also outperforms the Oracle method in the \reasoningII scenario, which has access to the test-domain labels. In this setting, we show that the \method routers benefit from further training with \methodplus, achieving 109.9\% of the dense expert's performance on average, demonstrating \method's ability to upcycle dense experts. We also show that \method allows seamless addition or removal of experts without requiring any additional finetuning to mitigate catastrophic forgetting.

One limitation of \method and \methodplus is that they require all experts models to have the same architecture because of the averaging of non-MLP parameters. 
Another limitation is that, although they do not require domain labels at inference time, they require at least some of them to be available in the training dataset. However, as we show in Appx.~\ref{sec:fewshot}, a few labeled samples are often sufficient: in the \reasoningI setting, for instance, \method retains over 99\% of the normalized average performance obtained with 820 samples when using only 20.
Future works may explore the extension of \method and \methodplus in scenarios with heterogeneous expert models.

\bibliography{arxiv2026} 
\bibliographystyle{plainnat}

\appendix
\section{Additional Related Work}
\label{app:relatedwork}

This section complements the related works proposed in \Secref{sec:intro}. 

\paragraph{Sparse and modular architectures.} Sparse architectures, such as MoEs \citep{shazeer2017outrageously, jacobs1991adaptive}, have emerged as a key technique for scaling LLMs with efficient compute. MoE models activate only a subset of experts per token, enabling extremely large parameter counts while controlling inference cost. Instruction-tuned MoE models \citep{shen2023mixture, zhu2025dynamic} have been shown to outperform dense counterparts with reduced FLOPs, highlighting the potential of sparse conditional computation in LLMs. Closely related to our work, DEMix \citep{gururangan2022demix} selectively mixes the feature maps of the experts based on domain labels. However, unlike our method, all experts are activated at inference time and mixed according to weights estimated from a validation set. Similarly, \citep{pfeiffer2022lifting, kudugunta2021beyond} propose multilingual and multi-task expert architectures with language-specific and task-conditioned routing.

\paragraph{Embarrassingly parallel training.} Efficient training of large models across heterogeneous corpora has motivated embarrassingly parallel frameworks that minimize costly synchronization. Branch-Train-Merge (BTM) \citep{li2022branchtrainmerge} decomposes LLM training into independent expert language models (ELMs), each trained on a domain subset. By branching from a seed model and ensembling, BTM improves in- and out-of-domain perplexity with minimal inter-worker communication. \citep{ersoy2025hdee} builds on BTM by exploring the effects of introducing heterogeneity in the number of training iterations and parameter count of the experts, finding that heterogeneous ensembles almost always achieve lower perplexities than homogeneous baselines. Branch-Train-MiX (BTX) \citep{sukhbaatar2024branch} integrates the BTM pipeline with MoE architectures: after expert training, BTX constructs a single MoE by integrating the experts' feedforward parameters as MoE experts and averaging the remaining parameters, then fine-tunes the MoE to learn routing. This approach retains embarrassingly parallel scalability during expert training while yielding unified sparse models with effective expert selection at inference. Finally, Branch-Train-Stitch (BTS) \citep{zhang2025bts} builds on parallel expert training by stitching domain experts into a generalist model via adding stitch layers that require further training steps.

\section{Pseudocode of \method}
\label{sec:pseudocode}
We present the pseudocode of \method in Algorithm ~\ref{alg:method}.

\begin{algorithm}[bt]
   \caption{\textbf{- \methodfull (\method)}}
   \label{alg:method}
    \begin{algorithmic}
        \STATE \textbf{Inputs:} Dense experts $\mathcal{E}_d$, $\forall d = 1, \dots, D$
        \STATE $\forall d = 1, \dots, D$, train the dense expert $f_d$ using the domain dataset $\mathcal{D}_d$. \\
        \STATE Perform the MoErging phase discussed in \Secref{sec:moerging}. \\
        \STATE Initialize: $A_l = 0_{H \times H}$, $b_l = 0_{H \times D}$ $\forall l = 1, \dots, L$. \\
        \FOR{\textbf{each} $d = 1, \dots, D$, \textit{in parallel}}
            \FOR{\textbf{each} $i = 1, \dots, B$, $l = 1, \dots, L$}
                \STATE $A_l \leftarrow A_l + F_{d,<l}(X_i)^\top F_{d,<l}(X_i)$ \\
                \STATE $b_l \leftarrow b_l + F_{d,<l}(X_i)^\top Y_i$ \\
            \ENDFOR
        \ENDFOR
        \FOR{\textbf{each} $l = 1, \dots, L$}
            \STATE $W^*_l = (A_l + \lambda I_h)^{-1} b_l$ \\
            \STATE $[W^*_l]_{:,c} = [W^*_l]_{:,c} / \norm{[W^*_l]_{:,c}}$ \\
        \ENDFOR
        \STATE \textbf{Return} MoE model $F$ with the parameters of the routers initialized using $W^*_l$, $\forall l = 1, \dots, L$
    \end{algorithmic}
\end{algorithm}

\section{Details on Experimental Setup}
\label{sec: exp_setup_appx}

\subsection{Details on the CLM experiments}

For \clmI experiments, we use the model and the domain datasets from \citep{ersoy2025hdee}, with the corresponding training hyperparameters. First, we pre-trained a seed 115M Llama model \citep{touvron2023llama} on the OpenWebText corpus dataset \citep{Gokaslan2019OpenWeb}, using a learning rate $=6 \cdot 10^{-4}$, with a cosine annealing scheduler with $1000$ warmup steps, a batch size $=16$ paragraphs of $1024$ padded tokens, with $4$ gradient accumulation steps and a total of $20000$ training iterations.
We then trained 5 experts, starting from the same pre-trained seed model, on the following M2D2 domains \citep{reid2022m2d2}: coding, mathematics, physics, history and events, and philosophy and thinking. We used a learning rate of $6 \cdot 10^{-5}$, a cosine annealing scheduler with $50$ warmup steps, a batch size of $16$ paragraphs of $1024$ padded tokens, $4$ gradient accumulation steps, and a total of $1000$ training iterations. 

Similarly, for the \clmII setting, we use a 1B Llama model \citep{touvron2023llama} pre-trained on the OpenWebText corpus dataset \citep{Gokaslan2019OpenWeb} for $30000$ iterations, with a learning rate $=4 \cdot 10^{-4}$, same scheduler as CLM1, and $2000$ warmup steps. We then trained 4 experts from the same pre-trained seed model, on the following domains: mathematics,  biology, legal and dialogue, for which we used the following datasets: OpenWebMath \citep{paster2023openwebmath}, peS2o, \citep{peS2o} Pile of Law \citep{hendersonkrass2022pileoflaw} and UltraChat 200k \citep{ding2023enhancing}. We used a learning rate of $4 \cdot 10^{-5}$, a cosine annealing scheduler with $100$ warmup steps, a batch size of $8$ paragraphs of $1024$ padded tokens, $4$ gradient accumulation steps, and a total of $2000$ training iterations.

For our \method experiments in the CLM settings, our default choice for the hyperparameters is $\lambda = 0.01$ and $k=1$. For BTX and \methodplus router training, we use a learning rate of $10^{-2}$ and $10^{-3}$ respectively in the CLM1 setting, and $10^{-2}$ for both BTX and \methodplus in the CLM2 setting. In both CLM1 and CLM2, we used $k=1$ as the default values for top-k routing and $100$ training iterations. Similar to \citep{sukhbaatar2024branch}, we also attempted to incorporate load balancing into BTX and \methodplus training, but its effects were negligible in our settings, so we decided not to use it. Finally, for our \methodplus experiments, we also attempted to modify the routers' softmax temperature, as motivated in \citep{fani2024accelerating}, but we observed no improvement and ultimately kept the temperature fixed at 1.

\subsection{Eval Datasets}
We specify the evaluation metrics for these datasets in \Tabref{tab:postdatasets}.

\begin{table}[t]
    \caption{Datasets used for evaluating the reasoning experiments, and their evaluation metrics.}
    \label{tab:postdatasets}
    \centering
    \begin{adjustbox}{width=\linewidth}
        \begin{tabular}{l p{7cm} p{5cm}}
            \toprule
            Dataset & Description & Metric \\
            \midrule
            HumanEval \citep{chen2021evaluating} 
            & Code synthesis benchmark consisting of programming problems with natural language descriptions and unit tests. 
            & pass@1: proportion of correctly solved problems. \\
            
            GSM8k \citep{cobbe2021training} 
            & Grade-school math reasoning dataset composed of linguistically diverse word problems requiring multi-step reasoning. 
            & Accuracy: proportion of problems for which the model predicts the correct final answer. \\
            
            M\_ARC \citep{lai2023okapi} 
            & Multilingual ARC challenge question-answering dataset used in Okapi evaluations, focusing on multiple-choice reasoning tasks. 
            & Accuracy: percentage of correct answers. \\
            
            IFEval \citep{zhou2023instruction} 
            & Instruction-Following Evaluation benchmark for large language models with prompts containing explicit, verifiable constraints. 
            & Task compliance rate: percentage of outputs satisfying all instructions. \\
            \bottomrule
        \end{tabular}
    \end{adjustbox}
\end{table}

\section{Causal language modeling experiments with frozen non-MLP domain experts}
\label{sec:dumefrozen}

\begin{table}[t]
    \caption{\clmI experiments where only MLP layers are trained in the original dense experts. The best and second best results, excluding the results of the experts in the corresponding domains and the Oracle baseline, which, unlike the other methods, uses the domain label at test time, are highlighted in \textbf{bold} and \underline{underlined}, respectively. 
    Normalization is performed using the performance of the full-finetuned experts in \Secref{sec:clmres} in their respective domains to highlight the differences from those experiments.}
    \label{tab:clmfrozen}
    \centering
    \begin{adjustbox}{width=\linewidth}
        \begin{tabular}{c|ccccc|c}
            \toprule
            \textbf{Method} & \textbf{Coding} & \textbf{Mathematics} & \textbf{Physics} & \textbf{History} & \textbf{Philosophy} & \textbf{Average} \\
            \midrule
            Coding expert     & \num{98.6}{4.0}  & \num{72.8}{5.6}  & \num{83.8}{3.5}  & \num{84.3}{1.0}  & \num{85.4}{0.8}  & \num{85.0}{3.0} \\
            Math expert       & \num{92.5}{3.7}  & \num{95.3}{7.5}  & \num{83.9}{3.6}  & \num{79.3}{0.9}  & \num{80.6}{0.7}  & \num{86.3}{3.3} \\
            Physics expert    & \num{85.7}{3.4}  & \num{65.3}{5.0}  & \num{98.6}{4.1}  & \num{84.6}{1.0}  & \num{85.6}{0.8}  & \num{84.0}{2.9} \\
            History expert    & \num{73.6}{3.0}  & \num{47.2}{3.6}  & \num{73.1}{3.1}  & \num{99.1}{1.2}  & \num{98.6}{0.9}  & \num{78.3}{2.4} \\
            Philosophy expert & \num{74.9}{3.1}  & \num{49.1}{3.8}  & \num{74.0}{3.1}  & \num{96.5}{1.2}  & \num{100.0}{0.9} & \num{78.9}{2.4} \\
            Model averaging   & \num{89.7}{3.7}  & \num{68.9}{5.3}  & \num{87.1}{3.7}  & \num{92.5}{1.0}  & \num{94.2}{0.9}  & \num{86.5}{2.9} \\
            \midrule
            Oracle            & \num{101.7}{1.3} & \num{100.5}{1.5} & \num{101.0}{0.0} & \num{99.7}{0.3}  & \num{100.4}{0.2} & \num{100.7}{0.5} \\
            Random routing    & \num{90.8}{0.8}  & \num{72.1}{0.9}  & \num{87.9}{0.9}  & \num{91.9}{0.3}  & \num{93.7}{0.5}  & \num{87.3}{0.1} \\
            BTX               & \num{95.1}{3.7}  & \num{93.8}{7.5}  & \num{94.1}{3.7}  & \num{97.8}{1.1}  & \num{98.3}{0.8}  & \num{95.8}{3.3} \\
            \rowcolor{gray!15}
            \textbf{\method (ours)}  & \bum{100.1}{1.5} & \bum{94.8}{1.4}  & \bum{98.0}{0.1}  & \bum{99.4}{0.3}  & \bum{100.0}{0.2} & \bum{98.5}{0.5} \\
            \rowcolor{gray!15}
            \textbf{\methodplus (ours)}  & \uum{95.8}{4.1}  & \uum{94.2}{7.5}  & \uum{95.7}{5.2}  & \uum{98.6}{1.4}  & \uum{98.8}{1.3}  & \uum{96.6}{3.9} \\
            \bottomrule
        \end{tabular}
    \end{adjustbox}
\end{table}

\begin{table}[t]
    \caption{\clmII experiments where only MLP layers are trained in the original dense experts. The best and second best results, excluding the results of the experts in the corresponding domains and the Oracle baseline, which, unlike the other methods, uses the domain label at test time, are highlighted in \textbf{bold} and \underline{underlined}, respectively. 
    Normalization is performed using the performance of the full-finetuned experts in \Secref{sec:clmres} in their respective domains to highlight the differences from those experiments.}
    \label{tab:clm2frozen}
    \centering
    \begin{tabular}{c|cccc|c}
        \toprule
        \textbf{Method} & \textbf{Mathematics} & \textbf{Biology} & \textbf{Legal} & \textbf{Dialogue} & \textbf{Average} \\
        \midrule
        Math expert        & \num{95.5}{4.1} & \num{90.6}{2.3} & \num{40.4}{3.3} & \num{90.5}{1.7} & \num{79.3}{1.8} \\
        Biology expert     & \num{70.8}{2.9} & \num{98.4}{2.5} & \num{29.0}{2.3} & \num{58.1}{10.4} & \num{64.1}{1.9} \\
        Legal expert       & \num{71.6}{3.1} & \num{86.5}{2.1} & \num{91.8}{7.2} & \num{89.1}{1.7} & \num{84.7}{2.4} \\
        Dialogue expert    & \num{66.9}{3.8} & \num{87.4}{3.4} & \num{33.1}{3.2} & \num{97.5}{3.4} & \num{71.2}{3.3} \\
        Model averaging    & \num{78.4}{3.6} & \num{92.4}{2.6} & \num{47.9}{3.9} & \num{91.9}{1.6} & \num{77.7}{2.1} \\
        \midrule
        Oracle             & \num{96.2}{3.9} & \num{98.3}{0.6} & \num{91.6}{6.7} & \num{95.8}{2.4} & \num{95.5}{2.7} \\
        Random routing     & \num{78.2}{4.2} & \num{91.6}{0.5} & \num{47.8}{4.2} & \num{91.9}{1.6} & \num{77.4}{1.6} \\
        BTX                & \bum{94.1}{2.7} & \uum{96.1}{2.5} & \num{89.9}{7.4} & \uum{94.3}{2.7} & \uum{93.6}{2.8} \\
        \rowcolor{gray!15}
        \textbf{\method (ours)}           & \num{91.7}{3.0} & \num{94.9}{0.9} & \uum{90.1}{9.8} & \num{93.7}{1.4} & \num{92.6}{3.1} \\
        \rowcolor{gray!15}
        \textbf{\methodplus (ours)}       & \uum{93.1}{4.6} & \bum{97.4}{1.5} & \bum{90.4}{7.3} & \bum{95.3}{1.3} & \bum{94.0}{3.0} \\
        \bottomrule
    \end{tabular}
\end{table}

\Cref{tab:clmfrozen,tab:clm2frozen} show the \clmI and \clmII experiments using domain experts that only trained the MLP layers, while leaving the other parameters frozen. Both \method and \methodplus excel in the \clmI setting, with \methodplus falling behind, underscoring how \method is again sufficient to achieve good performance in this scenario. In the \clmII setting with dense experts whose non-MLP layers are frozen, the \emph{domain-conditioned feature shift} appears more prominent, as methods that involve additional router training (\methodplus and BTX) achieve slightly better performance than \method. This may be due to a larger deviation of the MLP parameters from those of the seed model in this particular setting, and warrants further investigation in future work.

\section{Similarity of the fine-tuned parameters across domain experts}
\label{sec:cka}
In the MoErging phase (Section~\ref{sec:moerging}), we follow \citep{sukhbaatar2024branch} and average the non-MLP parameters of the dense experts, while keeping their MLP layers as separate MoE experts, since the latter are expected to encode most of the domain-specific knowledge. We check this assumption for our experts by measuring how similar the fine-tuned parameters of different domain experts are, separately for each group of modules. Specifically, we compute the CKA similarity between the fine-tuned parameters of each pair of domain experts, and average it over all expert pairs and layers.
Table~\ref{tab:cka} reports the results in the \clmII (Llama3-1B) and \reasoningI (Llama3-3B) settings.

In both settings, the non-MLP parameters are more similar across domain experts than the MLP ones ($+0.1666$ in \clmII and $+0.0839$ in \reasoningI), which supports merging the former and keeping the latter as separate experts, in line with the findings of \citep{sukhbaatar2024branch}.
Yet, it is worth to note that most of this gap is due to the embedding parameters, while the attention parameters are only slightly more similar across experts than the MLP ones, especially in \reasoningI ($+0.0059$).

\begin{table}[h]
\centering
\caption{Average CKA similarity between the fine-tuned parameters of different domain experts, per module group. Higher values indicate that a module group is more similar across domain experts.}
\label{tab:cka}
\small
\begin{tabular}{l|cc}
\toprule
Module group & \clmII (1B) & \reasoningI (3B) \\
\midrule
MLP            & 0.1246 & 0.0121 \\
Attention      & 0.1620 & 0.0180 \\
Embedding      & 0.2974 & 0.1473 \\
Non-MLP (all)  & 0.2912 & 0.0960 \\
\midrule
Non-MLP $-$ MLP   & $+0.1666$ & $+0.0839$ \\
Attention $-$ MLP & $+0.0375$ & $+0.0059$ \\
\bottomrule
\end{tabular}
\end{table}

\section{Computational costs}
\label{sec:compcosts}

Unlike the existing methods, which require additional training, our method only necessitates forwarding the domain datasets once, thereby dramatically reducing costs compared with further multi-task training. The computational costs of \Eqref{eq:rr-solution} are $\mathcal{O}(nH^2) + \mathcal{O}(H^3) + \mathcal{O}(nHC) + \mathcal{O}(H^2 C)$, which could be further optimized with techniques like Cholesky decomposition \citep{cholesky2005sur}. When we collect all the updates from \Eqref{eq:update-rule-general}, the total computational costs are the same as if we had a single batch containing all the inputs. In \method, the number of domains is negligible, and we need to substitute $n$ with $nT$. For simplicity, here we assume a constant number of tokens $T$ per data point. \method requires computing the ridge regression solution once for every MoE block. Therefore, the final computational costs are the forward costs of the inputs plus the cost of computing the ridge regression solutions, which is $\mathcal{O}(nTH^2 + H^3)$.
Note that, unlike BTX, our algorithm does not require backpropagation because it does not require additional training.

Specifically, as in \citep{fani2024accelerating, legate2023guiding}, we approximate backward passes to cost about twice as much as forward passes. Under this assumption, BTX incurs roughly three times the computational cost of \method for the same number of input tokens. In practice, however, BTX is likely more expensive because it typically requires more training samples. For example, while 80 samples may suffice for \method (as we show in \ref{sec:fewshot}), BTX needs to train the routers for several steps before converging, thus requiring more forward and backpropagation passes. Moreover, even though \method involves $L$ matrix inversions, these additional costs are negligible in practice, as the matrices involved are relatively small (recall that the size of the matrix that has to be inverted is the hidden size of the model, which is fixed). We also empirically confirmed that the matrix inversion costs are negligible compared to the forward and backpropagation costs in our experiments.

Finally, in Appendix \ref{sec:hyperparams}, we also show that a smaller token window per paragraph is sufficient to retain most of the performance, jointly reducing computational costs.

\section{Details on the merging algorithm}
\label{sec:merging}

In both the \clmI and \clmII settings, we compared \method with the following merging baselines:

\begin{itemize}
    \item \emph{Task arithmetic} \citep{ilharco2022editing, zhou2024metagpt} aggregates the dense experts with a weighted sum of the task vectors and the base model: $\theta = \theta_0 + \sum_{d=1}^D \lambda \tau_d$, where $\tau_d \coloneqq \theta_d - \theta_0$. We tested all $\lambda \in \{0.05, 0.1, 0.2, 0.3, 0.4, 0.5\}$ and found that the best value is $\lambda = 0.3$ for \clmI and $\lambda = 0.4$ for \clmII.
    \item \emph{TIES Merging} \citep{yadav2023ties} trims insignificant parameter updates with a top-k rule, resolves sign conflicts across models, and merges only parameters that agree in sign. We tested all $k \in \{\, 5, 10, 20, 50, 100\}$ and found that the best value is $k=50$ in \clmI and $k=20$ in \clmII.
    \item \emph{DARE} \citep{yu2024language} randomly filters out $p\%$ of the experts' parameters and then aggregates the remaining ones with a task arithmetic-style merging. For this baseline, we used the best performing $\lambda$ value from task arithmetic, and evaluated all $p \in \{0.05, 0.1, 0.2, 0.5, 0.8, 0.9\}$. For \clmI, we found that the best value is $p=0.05$, while for \clmII the best is $p=0.8$.
    \item \emph{Fisher Merging} \citep{matena2022merging}, which aggregates the dense experts based on the Fisher information matrix, which determines how much information an observable random variable carries about unknown parameters of a probabilistic model. To construct this matrix, we estimated the gradients with 100 or 1000 iterations, and found that a more accurate estimate (with 1000 iterations) yield the best performance in both \clmI and \clmII settings.
    \item \emph{Consensus Merging} \citep{wang2024localizing}, which retains only the parameters that are important for at least a given number of $k$ tasks. The importance of the parameters is evaluated by retaining only the parameters $i$ which satisfy $|\tau_d(i)| \geq \lambda_i |\Bar{\tau}(i) - \tau_d(i)|$, where $\Bar{\tau} = \sum_d \tau_d$. For this method, following the same practice in \citep{wang2024localizing}, we evaluated all the combinations of $k = \{1, 2, \dots, D\} \times \{0.2, 0.3, 0.4, 0.5, 0.6\}$, and found that the best values are $k=3$ and $\lambda=0.3$ for \clmI, and $k=2$ and $\lambda=0.5$ for \clmII.
\end{itemize}

\section{Recursive DUME experiments}
\label{sec:dumerec}

\begin{table}[t]
    \caption{\clmI recursive DUME experiments, both with dense domain experts where all the parameters have been trained and where only the MLP parameters have been trained.}
    \label{tab:dumerecclm}
    \centering
    \begin{adjustbox}{width=\linewidth}
        \begin{tabular}{cc|cccccc}
            \toprule
            \textbf{Method} & \textbf{Trained layers} & \textbf{Coding} & \textbf{Mathematics} & \textbf{Physics} & \textbf{History} & \textbf{Philosophy} & \textbf{Average} \\
            \midrule
            \method & all & 95.2 & 86.4 & 93.0 & 97.4 & 97.9 & 94.0 \\
            \method REC & all & 95.5 & 87.6 & 93.3 & 97.3 & 97.9 & 94.3 \\
            \midrule
            \method & MLP only & 91.9 & 83.0 & 90.2 & 97.7 & 98.0 & 92.2 \\
            \method REC & MLP only & 92.3 & 84.6 & 90.7 & 97.7 & 98.0 & 92.6 \\
            \bottomrule
        \end{tabular}
    \end{adjustbox}
\end{table}

\begin{table}[t]
    \caption{\clmII recursive DUME experiments, both with dense domain experts where all the parameters have been trained and where only the MLP parameters have been trained. Best and second best results are highlighted in \textbf{bold} and \underline{underlined}, respectively.}
    \label{tab:dumerecclm2}
    \centering
    \begin{tabular}{cc|ccccc}
        \toprule
        \textbf{Method} & \textbf{Trained layers} & \textbf{Mathematics} & \textbf{Biology} & \textbf{Legal} & \textbf{Dialogue} & \textbf{Average} \\
        \midrule
        \method & all & 93.7 & 95.1 & 86.8 & 93.4 & 92.2 \\
        \method REC & all & 92.7 & 94.3 & 88.5 & 93.4 & 92.2 \\
        \midrule
        \method & MLP only & 88.6 & 93.3 & 88.9 & 92.4 & 90.8 \\
        \method REC & MLP only & 93.3 & 96.2 & 83.1 & 95.0 & 91.9 \\
        \bottomrule
    \end{tabular}
\end{table}

\Cref{tab:dumerecclm,tab:dumerecclm2} compare \method with \method REC in both the \clmI and \clmII settings. As shown, the \emph{domain-conditioned feature shift} problem does not significantly impact final performance in either scenario, demonstrating that \method is a robust and cost-efficient approximation.

\paragraph{Magnitude of the domain-conditioned feature shift.}
To directly quantify the domain-conditioned feature shift (Section~\ref{sec:dume}), we compare the feature maps of \method and \method REC in the \clmII setting. For each of the MoE layers L1, L4, L8, L12, and L15, we fit a PCA on the feature vectors collected immediately before the router, using 64 samples from each of the four domains.
We then select 256 probe tokens from the tokenizer vocabulary and forward them through both models.
For each probe token and layer, we project the feature vectors of the two models onto the first 10 principal components of the corresponding PCA, and compute the cosine similarity between the two projections. Table~\ref{tab:feature_shift} reports the cosine similarity averaged over the probe tokens, together with the lower bound of the 95\% Student's $t$ confidence interval of the mean.

At L1, the cosine similarity is 1, since all the parameters preceding this router are identical in the two models (the first router is computed in the same way, see Section~\ref{sec:dumevariants}). The similarity then decreases monotonically with depth, confirming that the feature shift accumulates in deeper layers, as argued in Section~\ref{sec:dumereasoning} for the \reasoningII setting. However, the shift remains small: even at the last layer, the mean cosine similarity is 0.997, with a 95\% confidence interval lower bound of 0.996. This is consistent with the negligible performance differences between \method and \method REC in Tables~\ref{tab:dumerecclm} and~\ref{tab:dumerecclm2}, and supports \method as an accurate approximation of the exact ridge regression solution computed by \method REC.

\begin{table}[h]
\centering
\caption{Cosine similarity between the feature maps of \method and \method REC, extracted before the MoE routers and projected onto the first 10 principal components, in the \clmII setting. Values are averaged over 256 probe tokens; the second row reports the lower bound of the 95\% Student's $t$ confidence interval of the mean. Layers are 0-indexed.}
\label{tab:feature_shift}
\small
\begin{tabular}{l|ccccc}
\toprule
Layer & L1 & L4 & L8 & L12 & L15 \\
\midrule
Mean cosine similarity & 1.0000 & 0.9998 & 0.9992 & 0.9985 & 0.9970 \\
95\% CI lower bound    & 1.0000 & 0.9998 & 0.9991 & 0.9983 & 0.9964 \\
\bottomrule
\end{tabular}
\end{table}

\section{Out-of-distribution \method experiments}
\label{sec:ood}

\begin{table}[t]
    \caption{\reasoningI \method experiments with out-of-distribution test data w.r.t. the training data. Best results, excluding the Oracle, are in \textbf{bold}. Results surpassing the Oracle are in {\color{darkgreen} green}.}
    \label{tab:reasoningood}
    \centering
    \begin{tabular}{c|cccc|c}
        \toprule
        \textbf{Method} & \textbf{Mathematics} & \textbf{Multilingual} & \textbf{Coding} & \textbf{Instr. following} & \textbf{Average} \\
        \midrule
        Oracle                  & \num{97.0}{2.5} & \num{100.2}{0.0} & \num{97.9}{20.4} & \num{84.0}{1.6} & \num{94.8}{4.9} \\
        \midrule
        \method & \bum{92.0}{0.9} & \gbum{100.4}{0.0} & \gnum{108.1}{6.2} & \gbum{86.9}{0.7} & \gbum{96.8}{1.3} \\
        \method OOD & \num{80.8}{1.0} & \gbum{100.4}{0.0} & \gbum{110.7}{9.4} & \num{82.9}{0.4} & \num{93.7}{2.5} \\
        \bottomrule
    \end{tabular}
\end{table}

\Cref{tab:reasoningood} compares \method with \method OOD. For the \method OOD experiments, we extract the ridge regression statistics from some of the datasets where the experts were originally trained in \citep{he2025mergebench}: Magicoder \citep{wei2023magicoder} for coding, DART-Math \citep{tong2024dart} for mathematics, Aya \citep{singh2024aya} for multilingual understanding, and TULU-3 persona IF \citep{lambert2024tulu} for instruction following.

Results show that our method performs well even when the distribution of the datasets from which \method collects ridge regression statistics differs from the test distribution. Moreover, \method OOD approaches the Oracle average performance, demonstrating that our method can be applied to OOD settings.

\section{\method combined with merging methods}
\label{sec:mergingmethods}

\begin{table}[t]
    \caption{Comparison of model merging algorithms and \method where the non-MLP layers are merged with the corresponding merging algorithm in the \clmI setting. Best and second best results are highlighted in \textbf{bold} and \underline{underlined}, respectively.}
    \label{tab:clmmerging}
    \centering
    \begin{adjustbox}{width=\linewidth}
        \begin{tabular}{cc|ccccc|c}
            \toprule
            & \textbf{Merging algorithm} & \textbf{Coding} & \textbf{Mathematics} & \textbf{Physics} & \textbf{History} & \textbf{Philosophy} & \textbf{Average} \\
            \midrule
            \multirow{6}{*}{\rotatebox[origin=c]{90}{\makecell{\textbf{w/o \method}}}} 
            & Model averaging   & \num{90.5}{3.7}  & \num{69.2}{6.2}  & \num{86.2}{3.5}  & \num{92.5}{0.9}  & \num{94.2}{0.7}  & \num{86.5}{2.9} \\
            & Task Arithmetic   & \num{98.5}{2.0}  & \num{84.8}{1.0}  & \num{92.7}{2.3}  & \num{91.6}{1.3}  & \num{93.1}{1.3}  & \num{92.2}{1.6} \\
            & TIES Merging      & \num{100.7}{1.7} & \num{86.6}{0.7}  & \num{94.0}{2.2}  & \num{91.0}{1.5}  & \num{92.7}{1.6}  & \num{93.0}{1.5} \\
            & DARE              & \num{98.5}{2.0}  & \num{84.7}{1.0}  & \num{92.7}{2.3}  & \num{91.6}{1.3}  & \num{93.1}{1.3}  & \num{92.1}{1.6} \\
            & Fisher Merging    & \num{95.6}{3.9}  & \num{75.9}{5.9}  & \num{90.7}{3.8}  & \num{94.9}{1.1}  & \num{96.4}{0.9}  & \num{90.7}{3.1} \\
            & Consensus Merging & \num{98.5}{4.0}  & \num{80.7}{6.3}  & \num{92.7}{3.9}  & \num{83.6}{1.0}  & \num{85.2}{0.8}  & \num{88.1}{3.2} \\
            \midrule
            \multirow{6}{*}{\rotatebox[origin=c]{90}{\makecell{\textbf{w/ \method}}}} 
            & Model averaging   & \num{97.5}{2.0}  & \num{92.3}{5.1}  & \num{95.8}{2.4}  & \uum{98.9}{1.3}  & \uum{99.0}{0.9}  & \num{96.7}{2.4} \\
            & Task Arithmetic   & \uum{100.8}{4.1} & \uum{97.6}{7.6}  & \num{98.5}{4.2}  & \num{97.3}{1.1}  & \num{97.5}{0.9}  & \num{98.3}{3.6} \\
            & TIES Merging      & \bum{101.4}{4.1} & \bum{99.1}{7.7}  & \uum{97.3}{4.2}  & \num{97.6}{1.1}  & \num{97.9}{0.9}  & \bum{99.0}{3.6} \\
            & DARE              & \uum{100.8}{4.1} & \uum{97.6}{7.6}  & \bum{98.6}{4.2}  & \num{97.3}{1.1}  & \num{97.5}{0.9}  & \uum{98.3}{3.6} \\
            & Fisher Merging    & \num{100.4}{4.1} & \num{96.5}{7.5}  & \bum{98.6}{4.2}  & \bum{99.6}{1.2}  & \bum{99.6}{0.9}  & \bum{99.0}{3.6} \\
            & Consensus Merging & \num{100.2}{4.1} & \num{95.7}{7.4}  & \num{97.7}{4.1}  & \num{94.1}{1.1}  & \num{94.7}{0.9}  & \num{96.5}{3.5} \\
            \bottomrule
        \end{tabular}
    \end{adjustbox}
\end{table}

\begin{table}[t]
    \caption{Comparison of model merging algorithms and \method where the non-MLP layers are merged with the corresponding merging algorithm in the \clmII setting. Best and second best results are highlighted in \textbf{bold} and \underline{underlined}, respectively.}
    \label{tab:clm2merging}
    \centering
    \begin{tabular}{cc|cccc|c}
        \toprule
        & \textbf{Merging algorithm} & \textbf{Mathematics} & \textbf{Biology} & \textbf{Legal} & \textbf{Dialogue} & \textbf{Average} \\
        \midrule
        \multirow{6}{*}{\rotatebox[origin=c]{90}{\makecell{\textbf{w/o \method}}}} 
        & Model averaging    & \num{79.9}{0.8}  & \num{93.5}{2.0}  & \num{50.0}{0.6}  & \num{93.5}{1.9}  & \num{79.2}{0.7} \\
        & Task Arithmetic    & \num{83.4}{0.0}  & \num{93.9}{0.0}  & \num{54.8}{0.1}  & \num{94.4}{1.8}  & \num{81.6}{0.5} \\
        & TIES Merging       & \num{85.0}{0.9}  & \num{95.8}{1.9}  & \num{56.0}{0.7}  & \num{95.3}{2.1}  & \num{83.0}{0.8} \\
        & DARE               & \num{83.1}{0.8}  & \num{93.8}{2.0}  & \num{54.7}{0.6}  & \num{94.4}{1.8}  & \num{81.5}{0.7} \\
        & Fisher Merging     & \num{81.2}{0.8}  & \num{92.3}{1.9}  & \num{59.1}{0.7}  & \num{91.4}{1.7}  & \num{81.0}{0.7} \\
        & Consensus Merging  & \num{86.8}{1.0}  & \num{96.2}{1.9}  & \num{56.4}{0.8}  & \num{94.9}{1.8}  & \num{83.6}{0.8} \\
        \midrule
        \multirow{6}{*}{\rotatebox[origin=c]{90}{\makecell{\textbf{w/ \method}}}} 
        & Model averaging   & \uum{93.4}{0.6}  & \uum{97.3}{1.2}  & \num{86.1}{0.9}  & \bum{97.9}{2.1}  & \bum{93.7}{0.7} \\
        & Task Arithmetic   & \num{92.5}{0.6}  & \num{96.7}{2.2}  & \num{87.5}{1.8}  & \num{95.8}{2.1}  & \num{93.1}{0.6} \\
        & TIES Merging      & \uum{93.4}{0.6}  & \num{96.5}{0.7}  & \uum{87.6}{2.0}  & \num{95.8}{2.1}  & \uum{93.3}{0.4} \\
        & DARE              & \num{92.8}{1.7}  & \num{96.7}{0.6}  & \bum{87.7}{0.6}  & \uum{96.1}{1.9}  & \uum{93.3}{0.6} \\
        & Fisher Merging    & \num{89.6}{1.5}  & \num{96.2}{0.9}  & \num{80.4}{1.0}  & \num{95.6}{1.0}  & \num{90.4}{0.5} \\
        & Consensus Merging & \bum{94.3}{0.2}  & \bum{97.4}{0.9}  & \num{87.3}{0.5}  & \num{95.6}{1.9}  & \bum{93.7}{0.5} \\
        \bottomrule
    \end{tabular}
\end{table}

\method algorithm is designed for training-free model MoErging. For the non-MLP layers, following the practice in \citep{sukhbaatar2024branch}, we aggregate them via simple parameter averaging. However, in principle, they can be combined with any desired strategy. In \Cref{tab:clmmerging,tab:clm2merging}, we compare different merging strategies for the non-MLP layers in \method, for both the \clmI and \clmII settings. When using \method together with any merging algorithm, we used the same set of hyperparameters as described in Appx. \ref{sec:merging}, with the following exceptions:
\begin{itemize}
    \item For \method + TIES Merging, we set $k=10$ in \clmI and $k=50$ in \clmII.
    \item For \method + DARE, we set $p=0.5$ in \clmII.
    \item For \method + Fisher Merging, we used 100 iterations for the estimation of the gradients in \clmII.    
\end{itemize}

As it is possible to observe, \method combined with a given merging algorithm always performs better than the corresponding merging algorithm alone. In particular, the best-performing methods in \clmI are \method with TIES Merging and \method with Fisher Merging, while in \clmII the best-performing ones are \method with simple averaging or with Consensus Merging. Furthermore, no clear best merging algorithm emerges, underscoring that all of them largely suffer from destructive interference and that \method (as other MoErging strategies) helps mitigate it and, in the MLP layers, completely prevents it, as no merging occurs in these layers.

\section{Additional incremental \method experiments}
\label{sec:moreincr}

\begin{figure}[h]
    \centering
    \includegraphics[width=\linewidth]{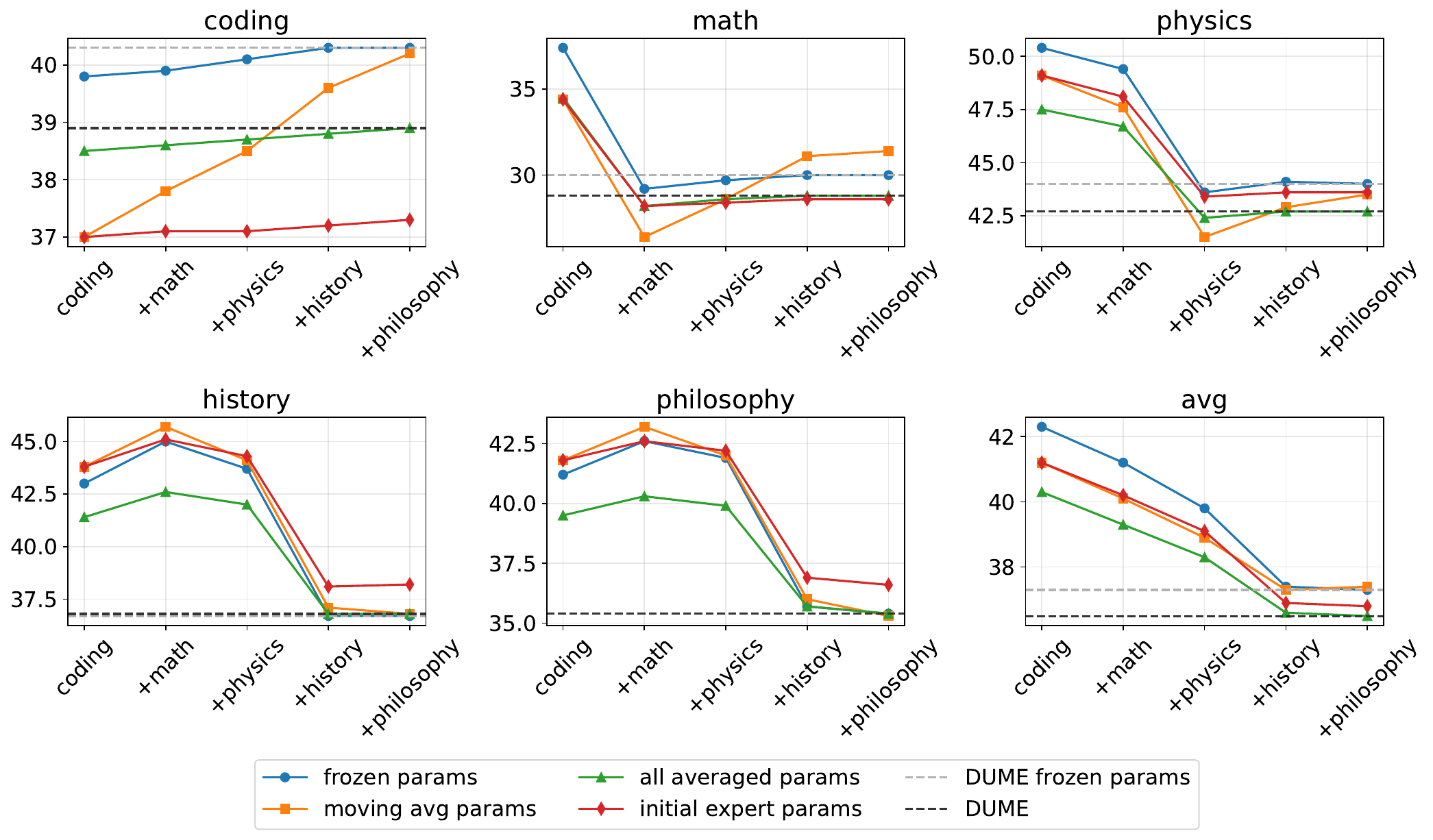}
    \caption{Dynamic incremental addition of new experts with \method in the \clmI setting for the \emph{frozen params} strategy.}
    \label{fig:incrclm}
\end{figure}

\Figref{fig:incrclm} shows the results for the \clmI setting when adding new experts sequentially, using the same strategies introduced in \Secref{sec:incrementalmain}. We observe the same trends as in the \clmII setting in the main paper. The sole difference we notice is that the \emph{moving avg params} strategy is more stable here, likely because the domains are more similar to each other and the models in this setting are simpler.

\begin{figure}[h]
    \centering
    \includegraphics[width=\linewidth]{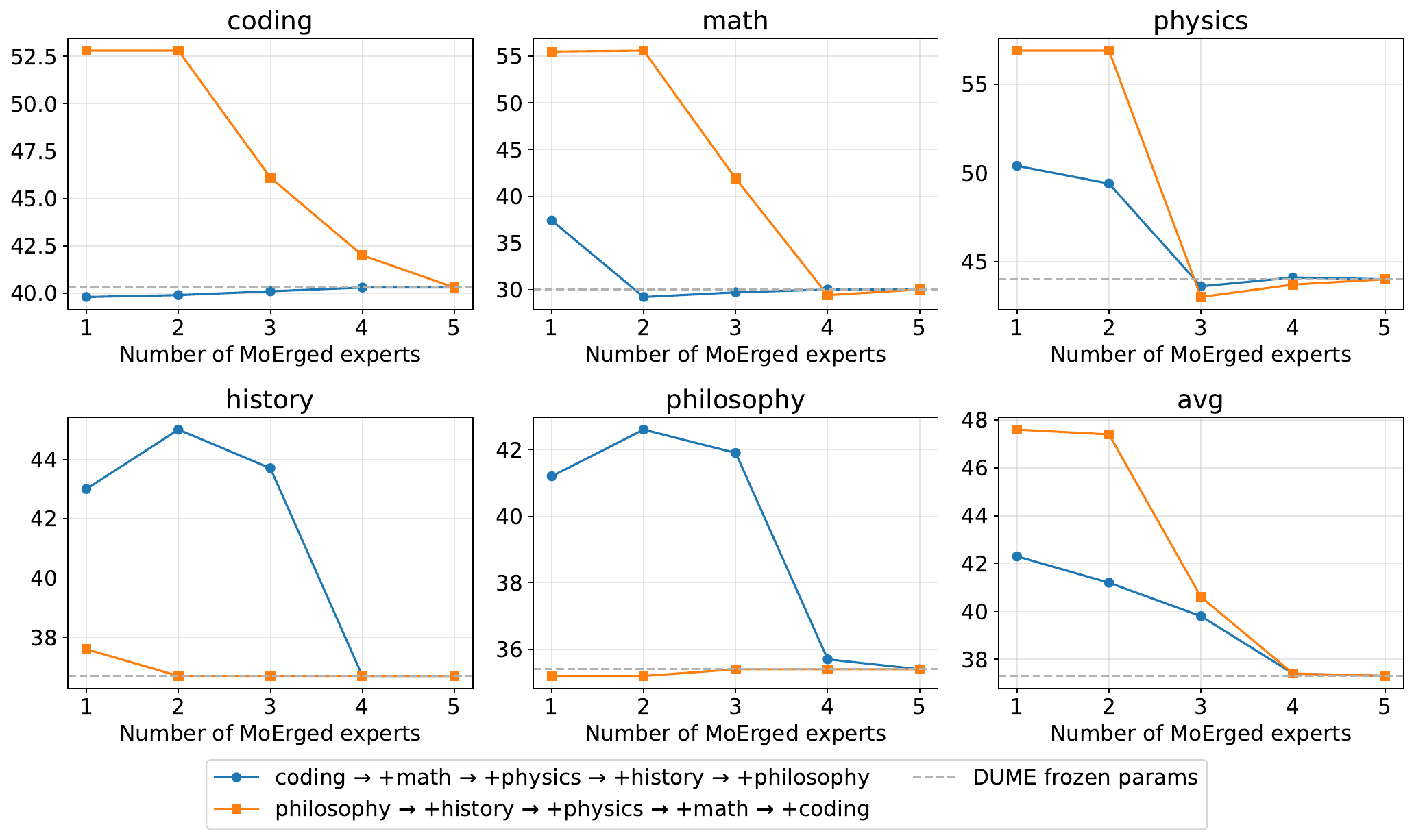}
    \caption{Comparison between two different orders of adding experts for \method in the \clmI setting for the \emph{frozen params} strategy.}
    \label{fig:difforderclm}
\end{figure}

\begin{figure}[h]
    \centering
    \includegraphics[width=\linewidth]{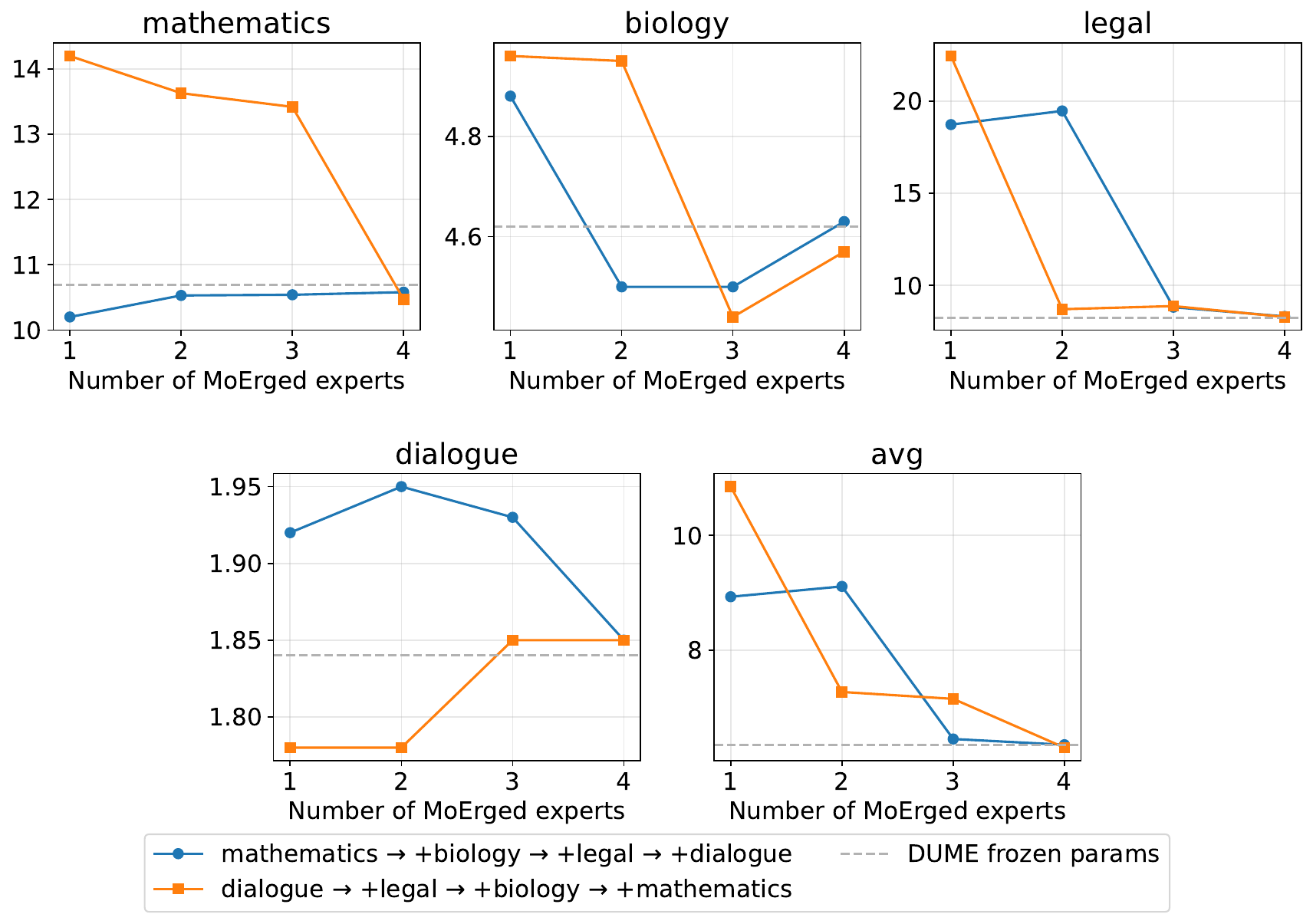}
    \caption{Comparison between two different orders of adding experts for \method in the \clmII setting.}
    \label{fig:difforderclm2}
\end{figure}

In \Cref{fig:difforderclm,fig:difforderclm2}, we show the results when adding experts using the \emph{frozen params} strategy in both the \clmI and \clmII settings, with two different orderings of the experts. As expected, given that \method is mathematically proven to be schedule-invariant with respect to the order in which the batch statistics are collected, and that with the \emph{frozen params} strategy, the non-MLP parameters are fixed, both curves converge to the final \method results across all plots.

\begin{figure}[h]
    \centering
    \includegraphics[width=\linewidth]{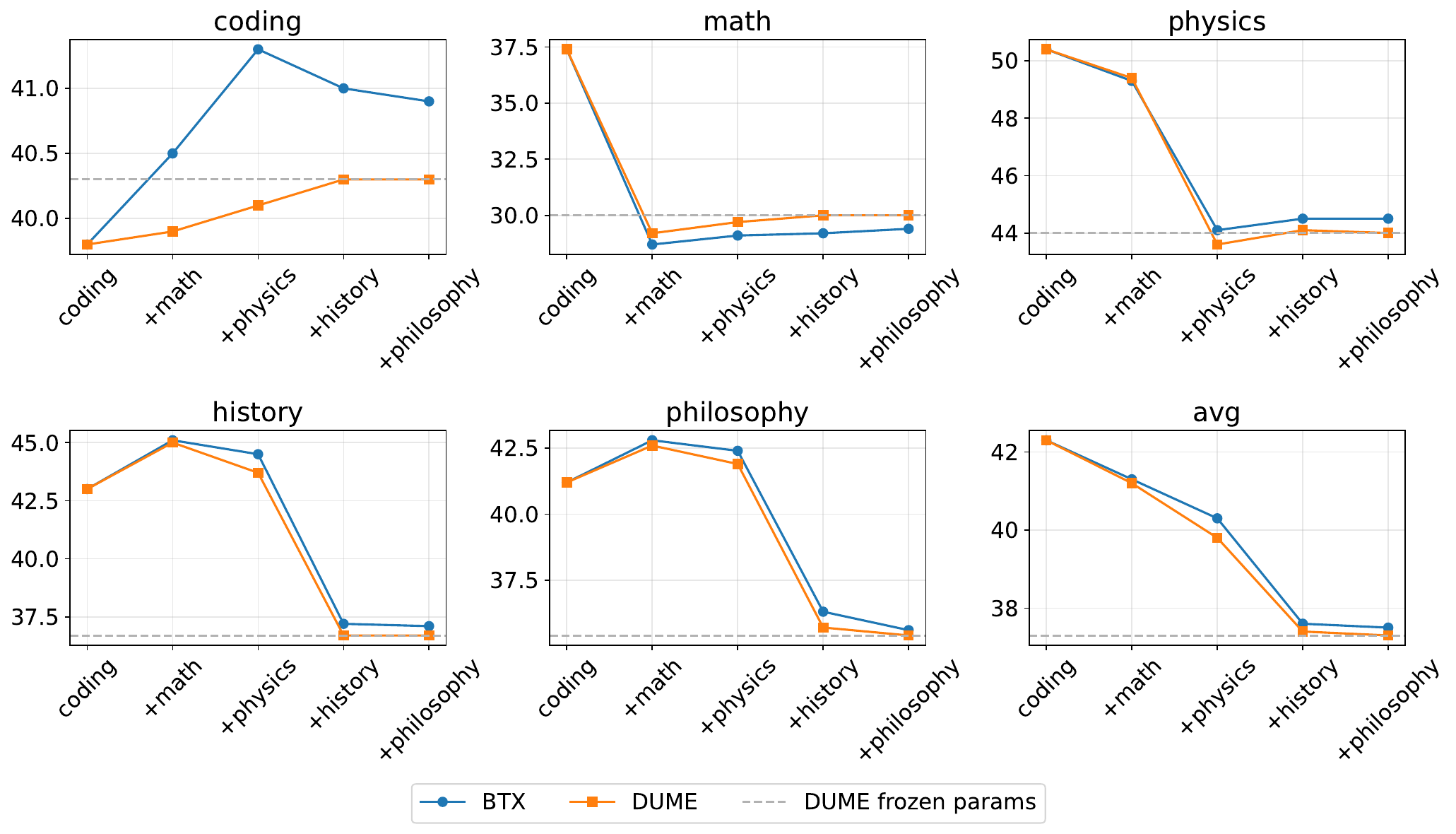}
    \caption{Comparison between \method and BTX when MoErging experts sequentially, using the \textit{frozen params} strategy, in the \clmI setting.}
    \label{fig:incrdumevsbtxclm}
\end{figure}

\begin{figure}[h]
    \centering
    \includegraphics[width=\linewidth]{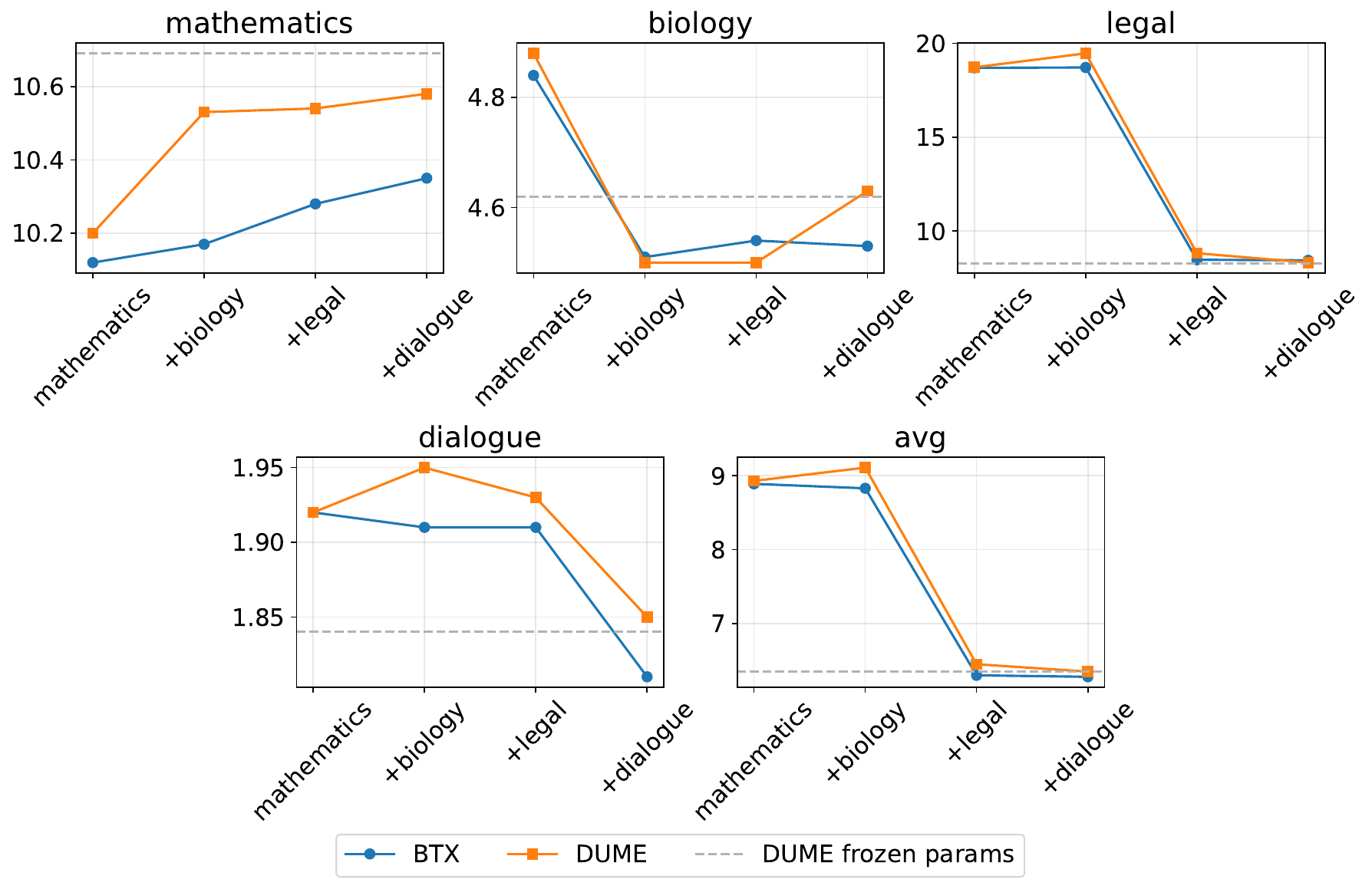}
    \caption{Comparison between \method and BTX when MoErging experts sequentially, using the \textit{frozen params} strategy, in the \clmII setting.}
    \label{fig:incrdumevsbtxclm2}
\end{figure}

Finally, \Cref{fig:incrdumevsbtxclm,fig:incrdumevsbtxclm2} show a comparison between \method and BTX when MoErging experts sequentially, using the \textit{frozen params} strategy, in the \clmI and \clmII settings, respectively. For BTX, when adding a new expert, we trained the routers for 50 iterations using samples from the new domain. Although BTX undergoes additional training and \method does not, \method perplexity is always comparable with BTX perplexity in both \clmI and \clmII settings, with \method even achieving a better average perplexity score in the \clmI setting.

\clearpage

\section{Dynamic subtraction of experts}
\label{sec:decr}

In \Secref{sec:dumedyn} we described how, in \method, the experts can also be removed from the pool in each MoE layer. In this section, we provide experiments showing that this is also possible in practice.

\begin{figure}[t]
    \centering
    \includegraphics[width=\linewidth]{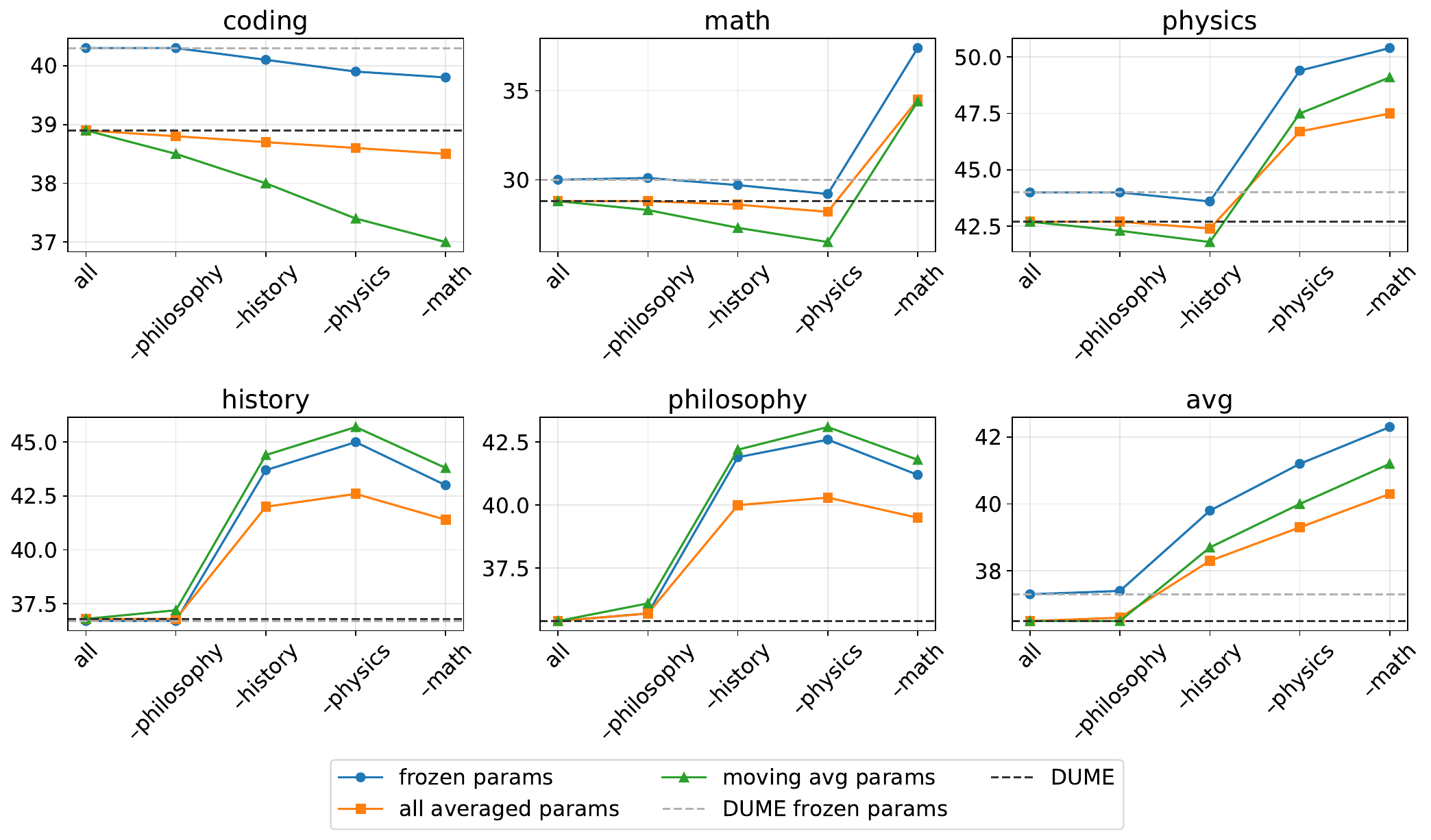}
    \caption{Dynamic decremental removal of experts with \method in the \clmI setting.}
    \label{fig:decrclm}
\end{figure}

\begin{figure}[t]
    \centering
    \includegraphics[width=\linewidth]{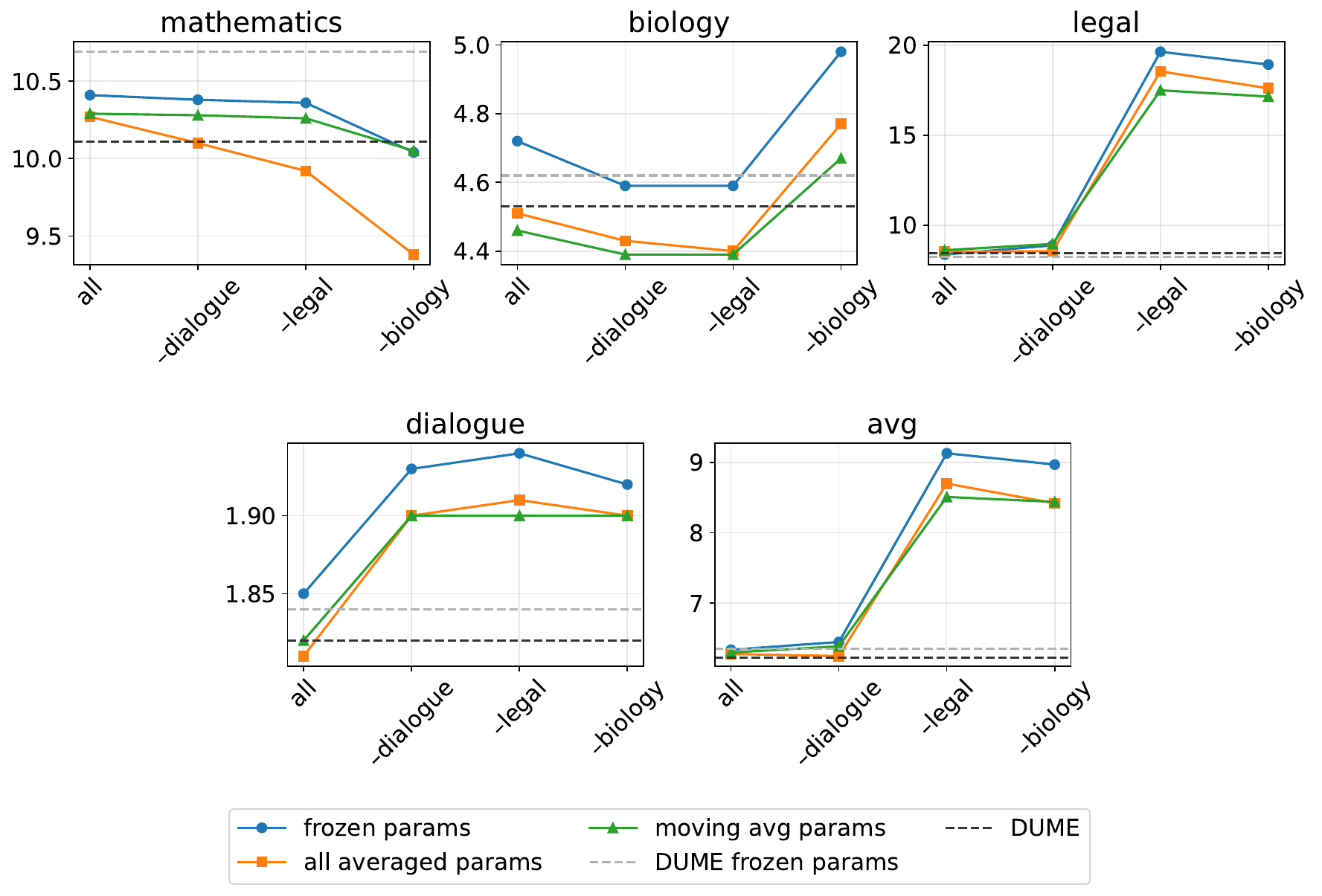}
    \caption{Dynamic decremental removal of experts with \method in the \clmII setting.}
    \label{fig:decrclm2}
\end{figure}

\Cref{fig:decrclm,fig:decrclm2} show the perplexity score for three different strategies when removing the experts as described in \Secref{sec:dumedyn}. As shown, when removing the experts, the \emph{moving avg params} strategy yields better results. However, in this scenario, \emph{all averaged params} is not an oracle method as the full average of the non-MLP parameters is available. Therefore, if achieving perfect machine unlearning is not needed, one should prefer this strategy, otherwise, \emph{moving avg params} is a valid alternative.

\begin{figure}[t]
    \centering
    \includegraphics[width=\linewidth]{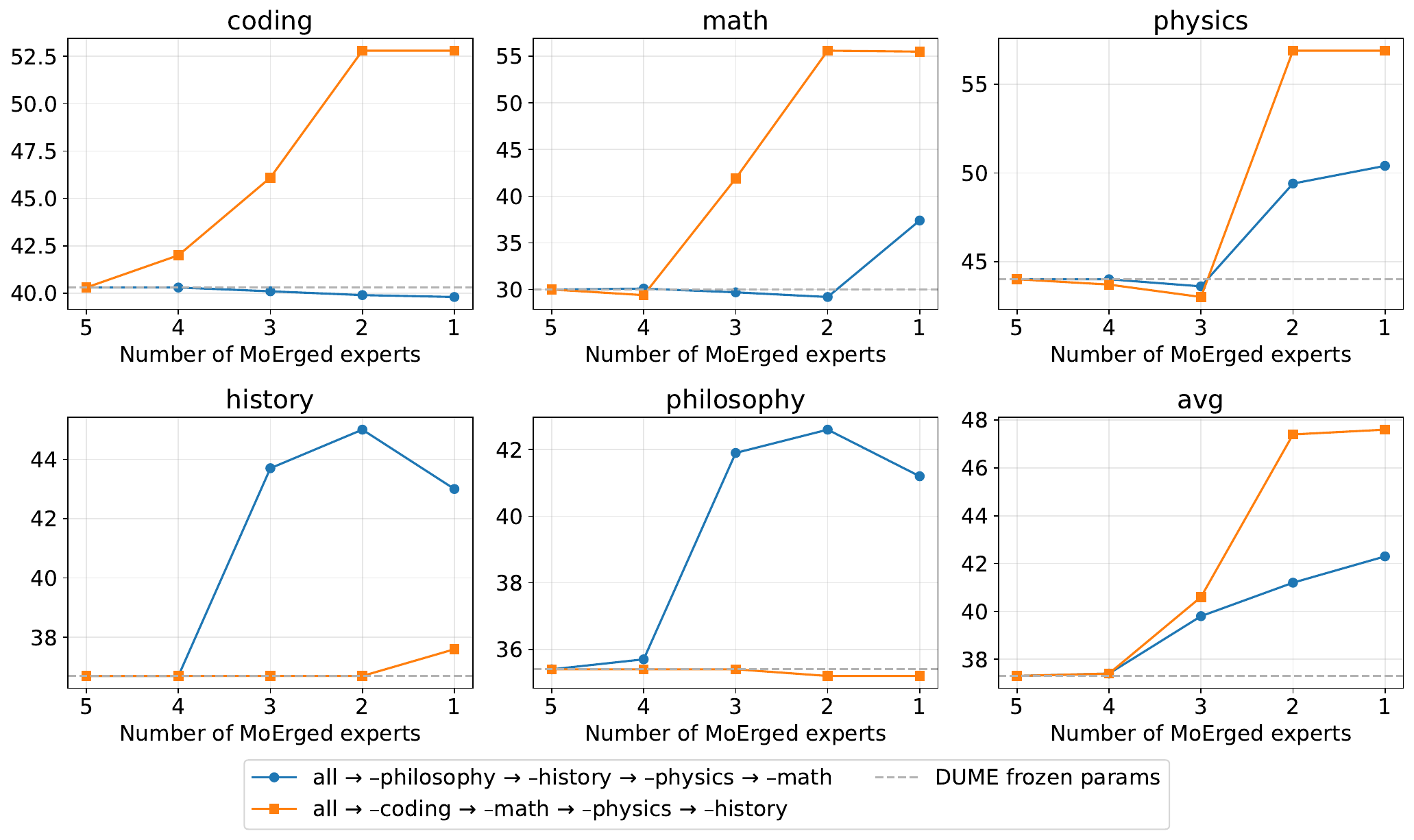}
    \caption{Comparison between two different orders of removing experts for \method in the \clmI setting.}
    \label{fig:difforderdecrclm}
\end{figure}

\begin{figure}[t]
    \centering
    \includegraphics[width=\linewidth]{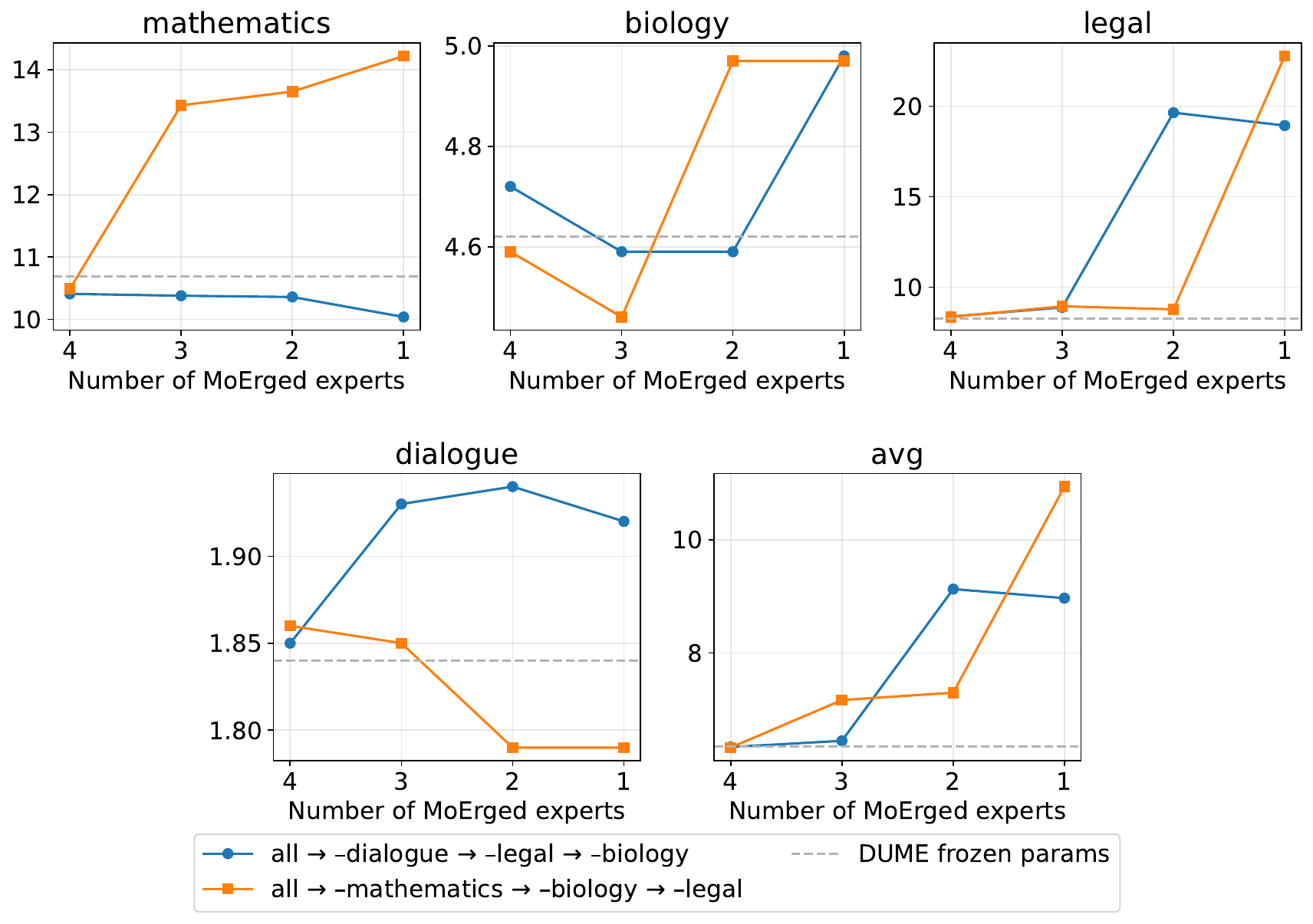}
    \caption{Comparison between two different orders of removing experts for \method in the \clmII setting.}
    \label{fig:difforderdecrclm2}
\end{figure}

Finally, \Cref{fig:difforderdecrclm,fig:difforderdecrclm2} show the results for the \textit{all averaged params} strategy of sequentially removing experts in two different orders. The two curves start from almost the same \method perplexity (tiny differences in the perplexities are likely due to numerical errors) and then diverge to the perplexity of the corresponding final remaining expert.

\section{Few-shot \method}
\label{sec:fewshot}

In all the \method experiments, we have limited the maximum number of samples used for each domain dataset to 16k for extracting the \method statistics. In many cases, this number is much larger than the actual number of samples used, since most datasets do not include such a large number of samples (\eg HumanEval \citep{chen2021evaluating} only counts 164 samples). Nevertheless, for some others, it is possible to reduce the number of samples to be used, and for all of them, we can also try very small amounts. In this section, we aim to investigate how \method reacts to being exposed to a few samples for which the domain of origin is known. This analysis is important for cases where a large pool of domain data, with the domain known, is unavailable and only a few labeled samples are available.

\begin{figure}[h]
    \centering
    \includegraphics[width=\linewidth]{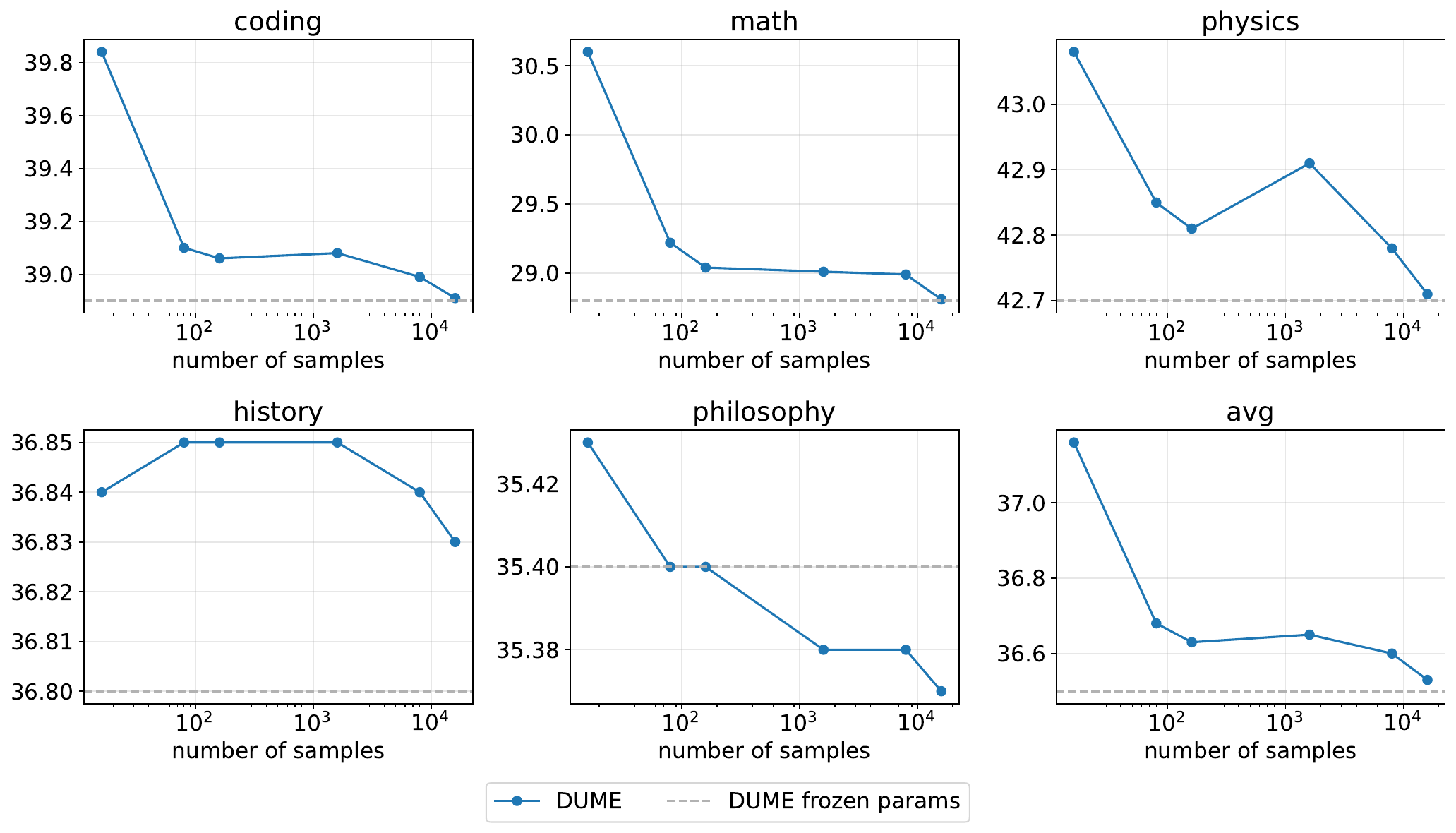}
    \caption{\method perplexity scores with various numbers of samples used to extract the ridge regression statistics, in the \clmI setting.}
    \label{fig:fewshotdumeclm}
\end{figure}

\begin{figure}[h]
    \centering
    \includegraphics[width=\linewidth]{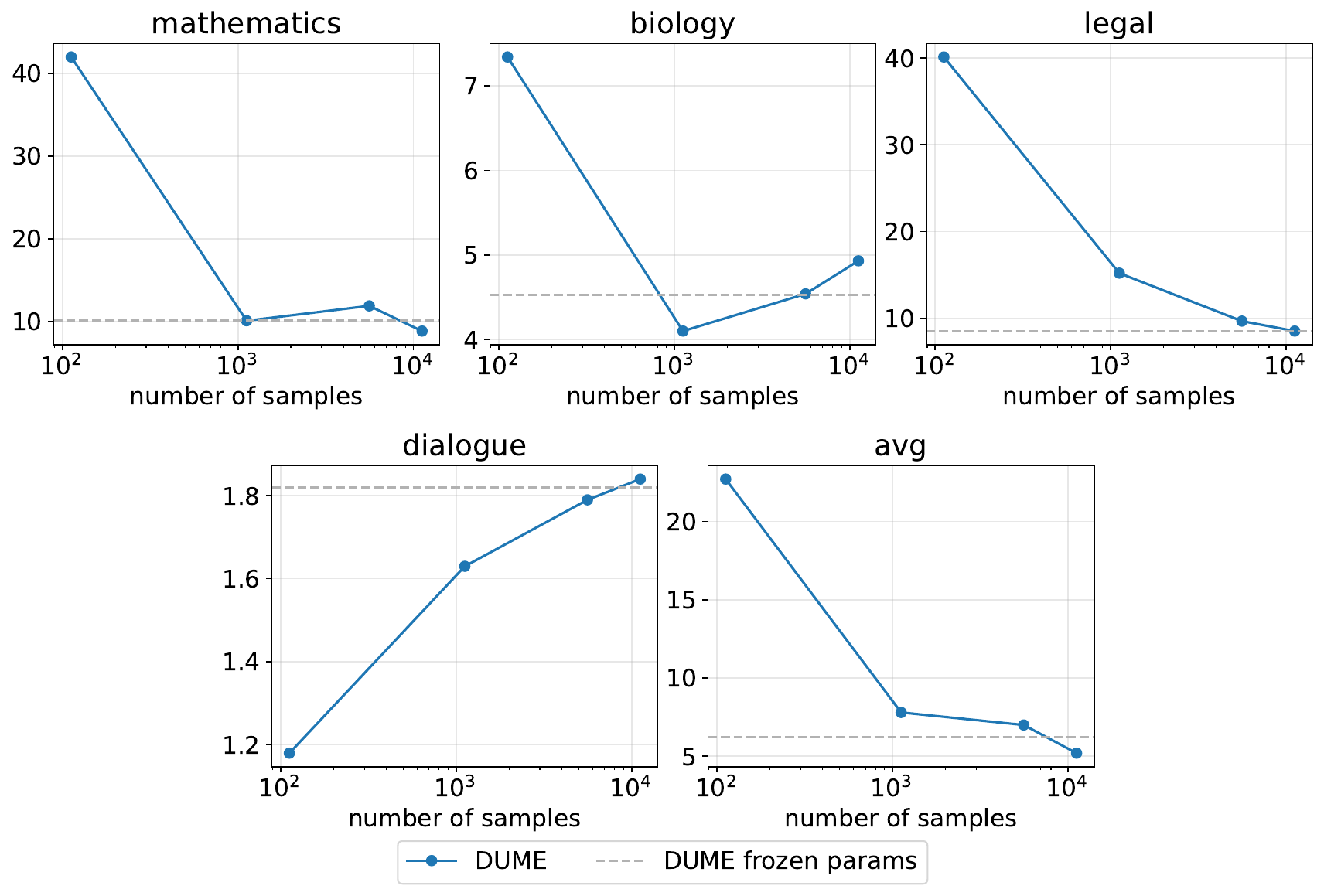}
    \caption{\method perplexity scores with various numbers of samples used to extract the ridge regression statistics, in the \clmII setting.}
    \label{fig:fewshotdumeclm2}
\end{figure}

\Cref{fig:fewshotdumeclm,fig:fewshotdumeclm2} show \method perplexity scores with various numbers of samples used to extract the ridge regression statistics, in the \clmI and \clmII settings, respectively. In particular, \method shows quite good performance already around ~100 samples in the \clmI domain and ~1000 samples in the \clmII domain, comparable with the performance of \method with the maximum amount of samples, showing that a large pool of labeled data is not mandatory to achieve good perplexities with \method.

Table~\ref{tab:fewshot_rmedium} extends this analysis to the \reasoningI setting, where we vary the number $n$ of samples per domain used to extract the ridge regression statistics. These experiments were run separately from those in Section~\ref{sec:dumereasoning} and use smaller evaluation sets. Hence, the absolute values differ from those of \method in Table~\ref{tab:reasoning}, and the different values of $n$ should only be compared within Table~\ref{tab:fewshot_rmedium}.

Saturation is domain-dependent. In Multilingual, the performance already saturates at $n=5$. In Coding and Instruction following, $n=20$ samples are already sufficient to surpass the performance obtained with $n=820$. In contrast, in Mathematics the performance increases monotonically with $n$ and does not appear to saturate even at $n=820$. On average, $n=20$ samples per domain yield a normalized performance of 95.0, close to the 95.7 obtained with $n=820$. Overall, these results confirm the findings of the CLM settings: a small number of labeled samples per domain is sufficient for \method to achieve near-optimal performance in most domains.

\begin{table}[h]
\centering
\caption{Normalized performance of \method in the \reasoningI setting with different numbers $n$ of samples per domain used to extract the ridge regression statistics. Note that these results come from a run separate from the one in Table~\ref{tab:reasoning}, with smaller evaluation sets.}
\label{tab:fewshot_rmedium}
\small
\begin{tabular}{c|cccc|c}
\toprule
$n$ per domain & Mathematics & Multilingual & Coding & Instr.\ following & Average \\
\midrule
5   & 43.9 & 102.6 & 117.1 & 64.3 & 82.0 \\
20  & 76.0 & 102.1 & 125.9 & 75.8 & 95.0 \\
100 & 84.8 & 103.2 & 146.4 & 64.3 & 99.7 \\
820 & 99.4 & 102.2 & 111.3 & 70.0 & 95.7 \\
\bottomrule
\end{tabular}
\end{table}

\clearpage

\section{On the hyperparameter choice}
\label{sec:hyperparams}

\begin{figure}[t]
    \centering
    \begin{subfigure}{.475\textwidth}
        \centering
        \includegraphics[width=\linewidth]{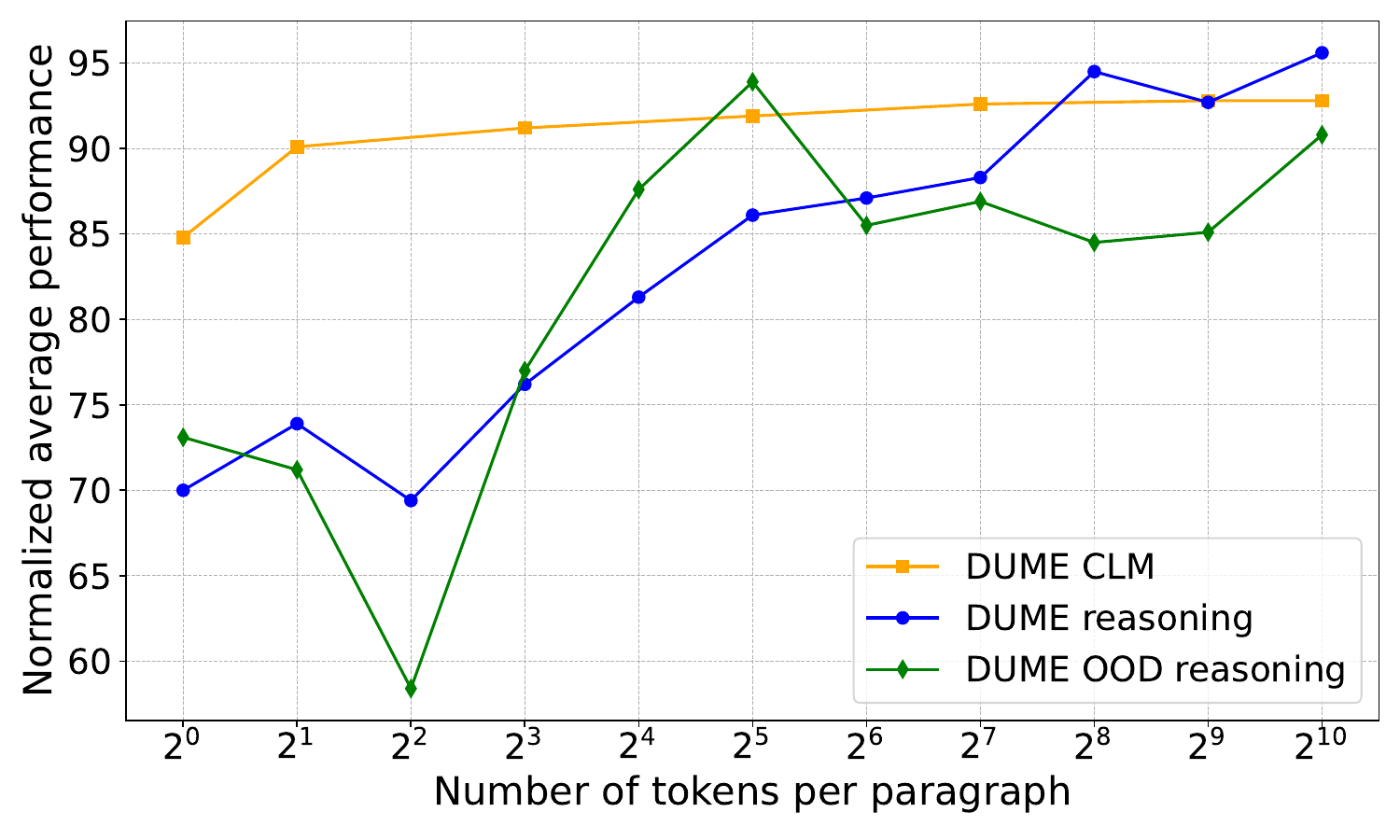}
        \caption{}
        \label{fig:hyperparams-a}
    \end{subfigure}\hfill
    \begin{subfigure}{.475\textwidth}
        \centering
        \includegraphics[width=\linewidth]{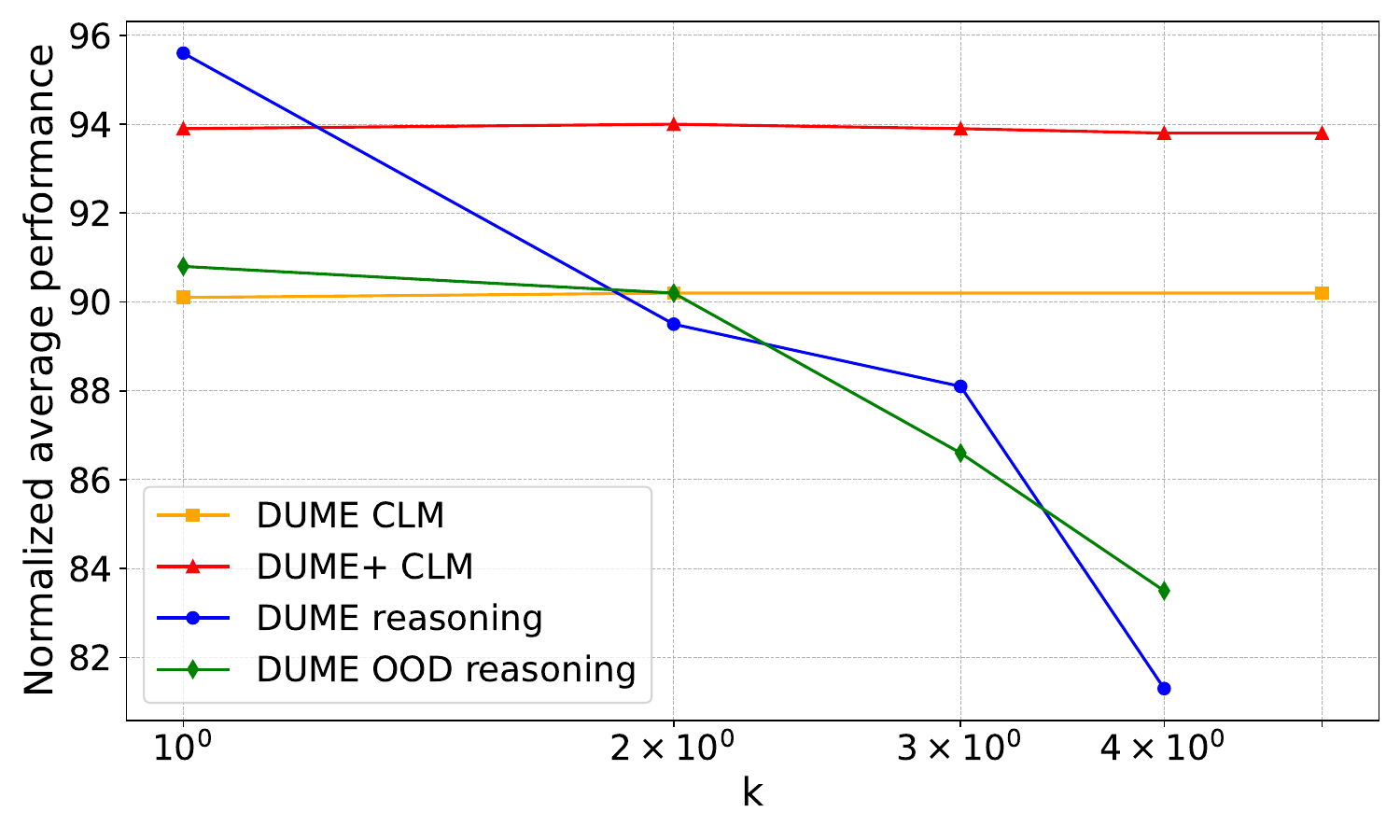}
        \caption{}
        \label{fig:hyperparams-b}
    \end{subfigure}\hfill
    \begin{subfigure}{.475\textwidth}
        \centering
        \includegraphics[width=\linewidth]{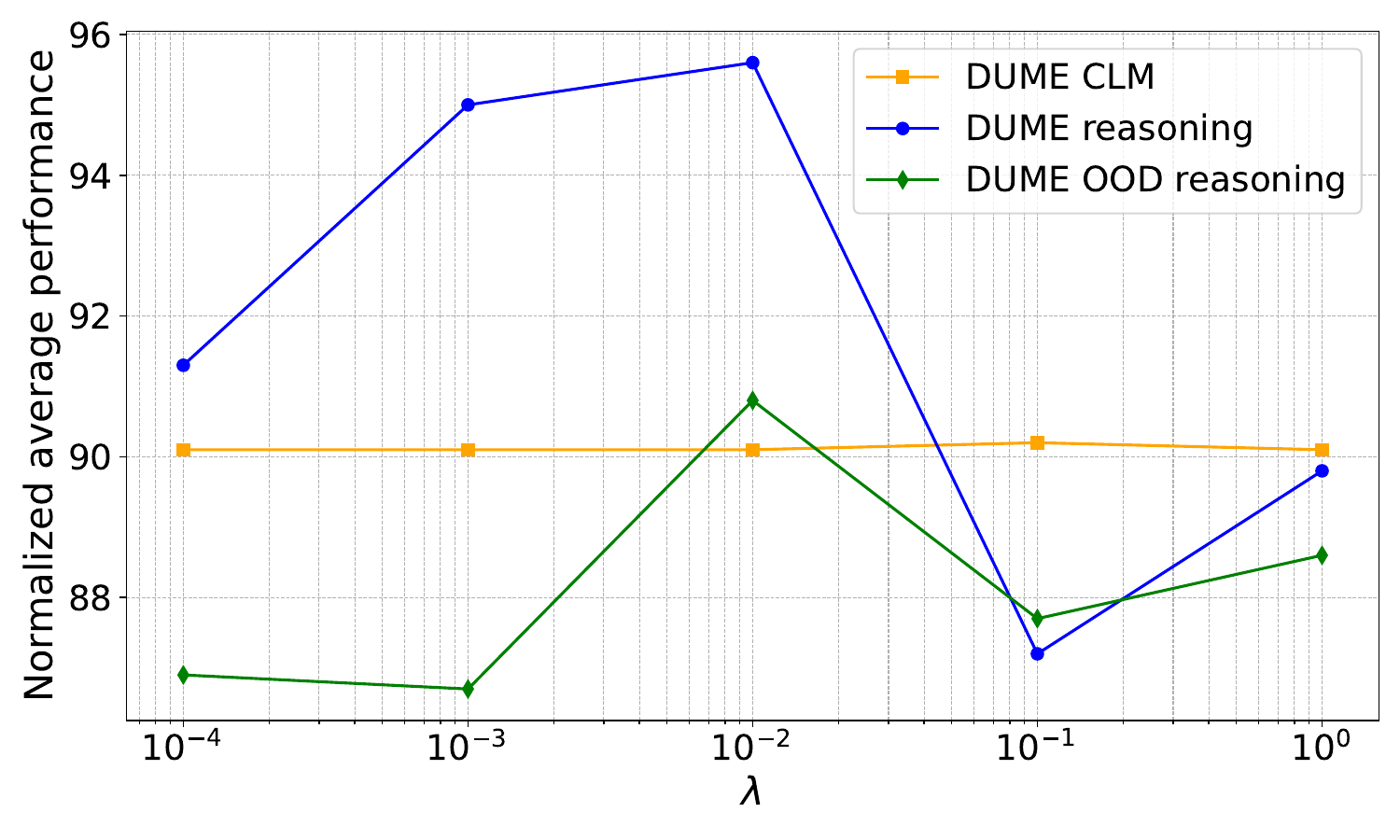}
        \caption{}
        \label{fig:hyperparams-c}
    \end{subfigure}
    \caption{(a) Normalized average performance in both CLM and reasoning settings, with various numbers of tokens used to extract the ridge regression statistics for each paragraph. (b) Normalized average performance in both CLM and reasoning settings, as $k$ of the top-k routing selecting increases. (c) Normalized average performance in both CLM and reasoning settings, with various values for the Tikhonov regularizer.}
    \label{fig:hyperparams}
\end{figure}

\Figref{fig:hyperparams} shows how the performance varies with the number of tokens used to extract the ridge regression statistics for each paragraph, $k$, and $\lambda$, in the \clmI and \reasoningI settings, for \method and \method OOD. In all experiments, we used the default hyperparameter values (except for the hyperparameter shown on the x-axis of these plots). The only exception is the \method CLM curve in \Figref{fig:hyperparams-b}, for which we used 2 as the number of tokens per paragraph, as it already achieved a sufficiently good performance.

As shown in \Figref{fig:hyperparams-a}, increasing the number of tokens generally yields improved results across all settings and methods. In the CLM setting, two tokens are sufficient to achieve 90\% of the dense experts' best performance, while 1024 tokens nearly reach the maximum score. These findings indicate that, within the CLM setting, computational costs can be further reduced since not all tokens in each paragraph are required. \Figref{fig:hyperparams-b} shows that \method is relatively unaffected by the choice of $k$, while in the reasoning setting, it is better to only activate the top-1 expert every time. This means that the forward and training costs of the final MoE model can be minimized, since using top-1 selection yields the same forward costs as the dense experts, provided we consider the forward through the routers negligible. Finally, \Figref{fig:hyperparams-c} shows how $\lambda = 0.01$ provides the best results in the reasoning setting, and that its choice is not important in the CLM setting, thus \method proves to be very robust to the choice of $\lambda$ in the CLM experiments.

\section{Additional comments on the results of the \reasoningII setting}
\label{sec:math}

We notice that the Math expert results are very poor in the Coding domain and quite poor in the other domains when compared with other domain experts. Additionally, we were surprised by how almost all the baselines surpass the Oracle in Mathematics, and the Mathematics expert itself.

\begin{figure}[t]
    \centering
    \begin{subfigure}[c]{0.48\textwidth}
        \centering
        \includegraphics[width=\linewidth]{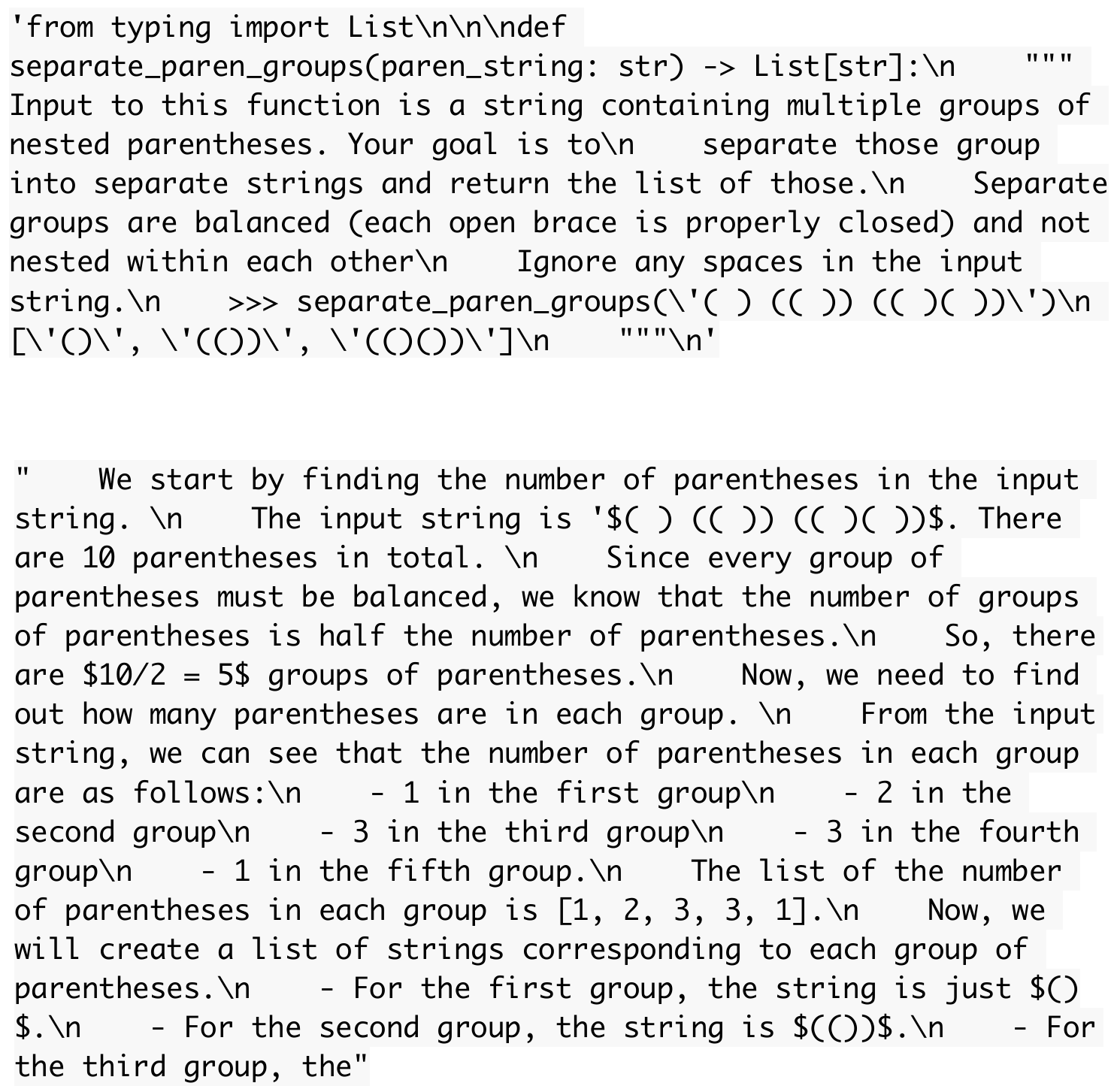}
        \caption{Completion in the coding domain. The math expert starts a lengthy line of reasoning instead of providing a valid code answer.}
        \label{fig:math1}
    \end{subfigure}
    \hfill
    \begin{minipage}[c]{0.48\textwidth}
        \centering
        \begin{subfigure}[c]{\linewidth}
            \centering
            \includegraphics[width=\linewidth]{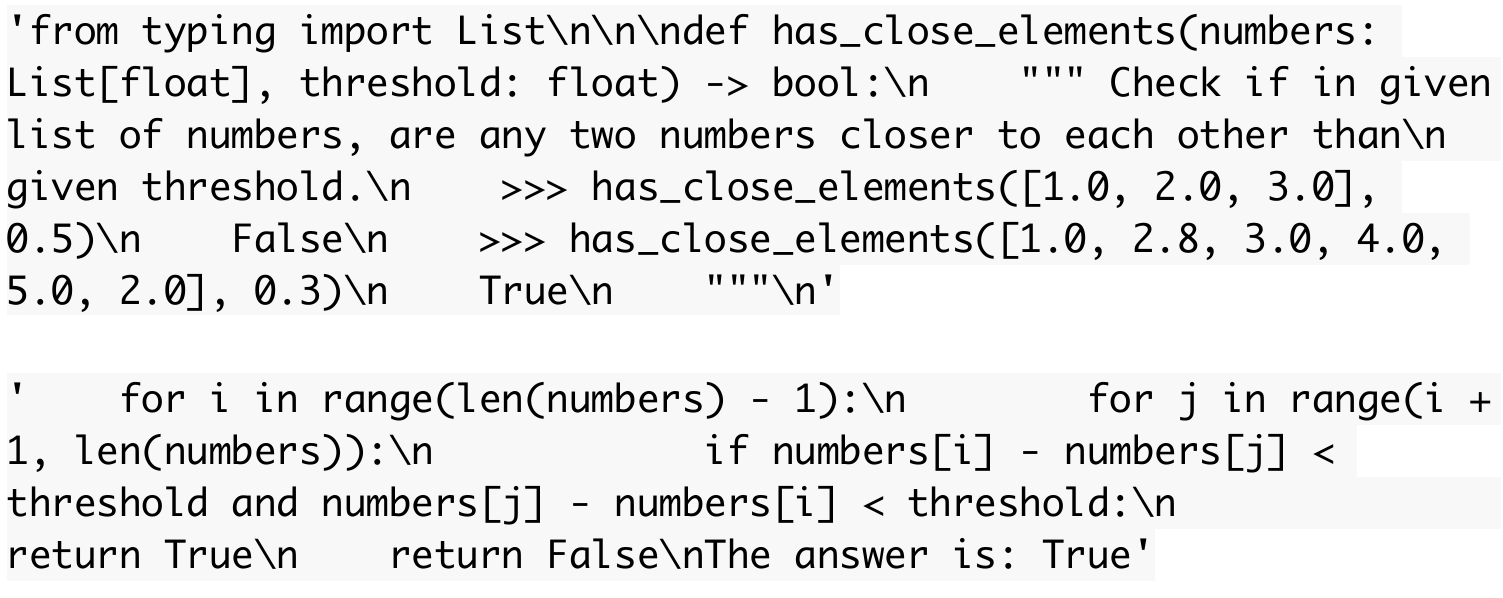}
            \caption{Completion in the coding domain. The math expert correctly generates the program, but then adds the text ``The answer is: True'' at the end, causing the code to fail.}
            \label{fig:topright}
        \end{subfigure}
        \vspace{0.5cm}
        \begin{subfigure}[c]{\linewidth}
            \centering
            \includegraphics[width=\linewidth]{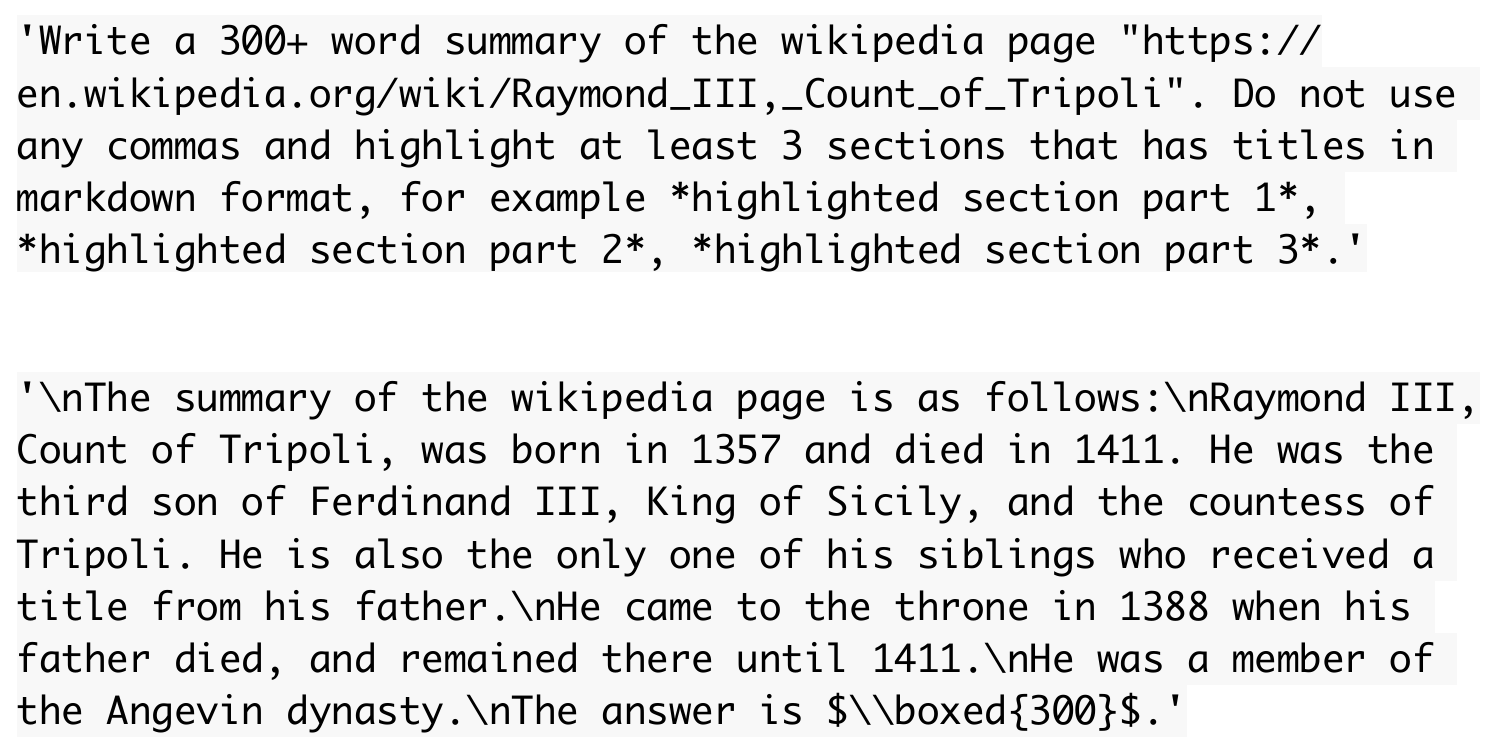}
            \caption{Completion in the instruction domain. Again, the math experts add an answer at the end, even if the request was only to write a summary and no final answers were requested.}
            \label{fig:bottomright}
        \end{subfigure}
    \end{minipage}
    \caption{Examples of completions by the math expert.}
    \label{fig:math}
\end{figure}

In \Figref{fig:math}, we show three examples of completions from the Math expert, indicating that this behavior stems from the Math expert likely overfitting the math datasets used in \citep{he2025mergebench} during training. Indeed, the math expert often fails to provide an answer that is correctly structured, although their reasoning abilities appear intact. This also explains why, when combined with other models, it largely improves results in Math, since the other models provide the capability to produce structured answers.

\section{Comparison between \method and a direct training of a softmax classifier}
\label{sec:dumevssoftmax}

\begin{table}[t]
    \caption{\method compared with a softmax classifier trained with RMSE loss in the \clmII setting.}
    \label{tab:dumevssoftmax}
    \centering
    \begin{tabular}{cccccc}
        \toprule
        \textbf{Method} & \textbf{Mathematics} & \textbf{Biology} & \textbf{Legal} & \textbf{Dialogue} & \textbf{Geometric mean} \\
        \midrule
        \method & 13.629 & \textbf{3.785} & \textbf{39.274} & \textbf{2.821} & \textbf{8.695} \\
        Softmax, no bias & \textbf{13.546} & 3.792 & 42.329 & 2.822 & 8.850 \\
        Softmax, + bias & \textbf{13.546} & 3.792 & 42.331 & 2.821 & 8.850 \\
        \bottomrule
    \end{tabular}
\end{table}

As an alternative to the routers constructed with \method, we fitted a softmax classifier trained with the RMSE loss to each router using AdamW optimization, using the original prompts' domains as the supervised signal, in the \clmII setting. \Tabref{tab:dumevssoftmax} shows the perplexity scores (the lower, the better) for three routers, for the whole 1024 context length (including the possible padding tokens). Since the \method router includes an intercept term, the with-bias variant was also trained to ensure a fair comparison.

The results of the two softmax experiments were identical to two decimal places, indicating negligible differences between including and excluding the bias term. \method outperforms both softmax routers across all domains except Math, where its performance is only marginally lower. Moreover, unlike softmax routers, \method allows for the dynamic addition and removal of experts without reconstructing the router from scratch. These findings align with expectations, as negligible domain-conditioned feature-shift, empirically demonstrated through the \method REC experiments presented in \Tabref{tab:dumerecclm2}, allows \method to approximate the optimal RMSE solution achieved by the closed-form ridge regression formulation.

\section{Evaluation of DUME routers' domain classification accuracy across different MoE layer depths}
\label{sec:routingacc}

\begin{table}[h]
    \caption{Classification accuracy across different layers.}
    \label{tab:layeraccuracy}
    \centering
    \begin{tabular}{lccccc}
        \toprule
        \textbf{Layer} & \textbf{Math} & \textbf{Biology} & \textbf{Legal} & \textbf{Dialogue} & \textbf{Balanced acc.} \\
        \midrule
        first (L0) & 91.8 & 88.1 & 96.7 & 77.0 & 88.4 \\
        L1 & 86.5 & 66.3 & 96.4 & 73.5 & 80.7 \\
        L2 & 85.5 & 54.3 & 96.3 & 78.5 & 78.6 \\
        L3 & 86.0 & 53.6 & 96.6 & 78.5 & 78.7 \\
        25\% (L4) & 85.6 & 53.1 & 96.8 & 80.3 & 78.9 \\
        L5 & 85.3 & 50.0 & 96.9 & 81.4 & 78.4 \\
        L6 & 86.3 & 48.8 & 96.9 & 82.4 & 78.6 \\
        L7 & 85.6 & 38.8 & 97.3 & 85.3 & 76.7 \\
        50\% (L8) & 86.7 & 38.3 & 97.4 & 86.2 & 77.2 \\
        L9 & 88.3 & 43.5 & 97.7 & 87.1 & 79.1 \\
        L10 & 89.6 & 47.0 & 97.7 & 87.3 & 80.4 \\
        L11 & 90.2 & 50.6 & 97.8 & 87.4 & 81.5 \\
        75\% (L12) & 90.9 & 52.9 & 98.0 & 88.1 & 82.4 \\
        L13 & 90.9 & 53.7 & 98.1 & 88.4 & 82.8 \\
        L14 & 91.2 & 55.2 & 98.0 & 88.9 & 83.3 \\
        final (L15) & 91.2 & 56.2 & 98.0 & 89.0 & 83.6 \\
        \bottomrule
    \end{tabular}
\end{table}

\Tabref{tab:layeraccuracy} shows the routing accuracy for each domain and the overall balanced routing accuracy of \method in the \clmII setting.

The routing accuracy significantly exceeds the random routing lower threshold (25\% routing accuracy for each domain), with values ranging from approximately 77 to approximately 88 depending on the specific layer for the balanced accuracy. This result highlights the effectiveness of the \method routers. Notably, both the balanced routing accuracy and the domain routing accuracy for the Math and Biology domains exhibit a U-shaped pattern, with the lowest accuracy observed in the initial to mid layers and the highest accuracy in the first and final layers. In contrast, the Legal and Dialogue domains demonstrate a more uniform accuracy distribution across layers.

\section{Experiments with larger router-training steps and token budgets for BTX}
\label{sec:btxsteps}

\begin{table}[h]
    \caption{BTX performance for different numbers of training steps in the \reasoningI setting. Note that these experiments were run separately from those in Table~\ref{tab:reasoning} and use smaller evaluation sets.}
    \label{tab:btxsteps}
    \centering
    \begin{tabular}{lccccc}
        \toprule
        \textbf{Num steps} & \textbf{Math} & \textbf{Multilingual} & \textbf{Coding} & \textbf{Instruction} & \textbf{Average} \\
        \midrule
        0 (random init) & 33.7 & \textbf{99.8} & 87.5 & 51.4 & 68.1 \\
        100 & 38.6 & \textbf{99.8} & \textbf{98.2} & \textbf{55.3} & \textbf{73.0} \\
        200 & 37.3 & 99.6 & 92.9 & 53.6 & 70.8 \\
        500 & 38.4 & 99.4 & 83.9 & 53.9 & 68.9 \\
        1000 & \textbf{41.7} & 99.7 & 82.1 & 54.7 & 69.6 \\
        \bottomrule
    \end{tabular}
\end{table}

To capture the intermediate normalized performance, in this section we evaluate BTX in the \reasoningI setting with different numbers of training steps. Results are shown in \Tabref{tab:btxsteps}. 
Note that these experiments were run separately from those in Section~\ref{sec:dumereasoning} and use smaller evaluation sets; hence, the results differ from those of BTX in Table~\ref{tab:reasoning} and should only be compared within Table~\ref{tab:btxsteps}.

Optimal performance is observed after 100 training steps, with subsequent training leading to decreased results, likely due to overfitting. The coding and instruction datasets are notably small (541 samples for instruction and 164 for coding), whereas the larger multilingual and math datasets maintain consistently high or improving performance. These findings further highlight two other advantages of \method: by employing the closed-form ridge regression solution, it 1) eliminates the need to find the optimal number of training steps for the routers and 2) demonstrates reduced sensitivity to dataset size imbalance.

\FloatBarrier

\end{document}